\documentclass[letterpaper,twocolumn,fleqn]{article} 

\usepackage{ist}
\usepackage{algorithm2e}

\usepackage[table]{xcolor}

\usepackage{graphicx}
\usepackage{subcaption}
\usepackage{float}

\title{Deep Learning Methods for Event Verification and Image Repurposing Detection}

\author{ M.Goebel*; University of California, Santa Barbara; Santa Barbara, CA\\
A. Flenner*; NAVAIR; China Lake, CA\\
L. Nataraj, B.S. Manjunath; Mayachitra Inc.; Santa Barbara, CA\\
\textit{`*'-equal contribution}}

\date{} 

\begin{document} 

\maketitle 

\thispagestyle{empty} 


\begin{abstract}

The authenticity of images posted on social media is an issue of growing concern. Many algorithms have been developed to detect manipulated images, but few have investigated the ability of deep neural network based approaches to verify the authenticity of image labels, such as event names. In this paper, we propose several novel methods to predict if an image was captured at one of several noteworthy events. We use a set of images from several recorded events such as storms, marathons, protests, and other large public gatherings. Two strategies of applying pre-trained Imagenet network for event verification are presented, with two modifications for each strategy. The first method uses the features from the last convolutional layer of a pre-trained network as input to a classifier.  We also consider the effects of tuning the convolutional weights of the pre-trained network to improve classification.  The second method combines many features extracted from smaller scales and uses the output of a pre-trained network as the input to a second classifier. For both methods, we investigated several different classifiers and tested many different pre-trained networks.  
Our experiments demonstrate both these approaches are effective for event verification and image re-purposing detection.
The classification at the global scale tends to marginally outperform our tested local methods and fine tuning the network further improves the results.  
\end{abstract}


\section{Introduction}
\label{sec:intro}

Today social media websites are emerging as a dominant news source, but verifying the validity of the news stories is a difficult problem. 
In the few months before the 2016 US Presidential elections, the average American saw at least one fake news story, and of those who saw one, half of them believed it to be true~\cite{allcott2017social}.
In practice, countering these sources of fake news is a complex problem. 
There is little entry cost to distributing false information on Facebook, with large potential for ad revenue if the story gains popularity.  While in the past people relied on a few well known sources, the website publishing false stories is often removed before being identified as illegitimate~\cite{allcott2017social}.
For these reasons, an automated algorithm is needed to identify these fake stories before they reach a large number of users.

\begin{figure*}[ht]
\centering
\includegraphics[width=0.95\linewidth]{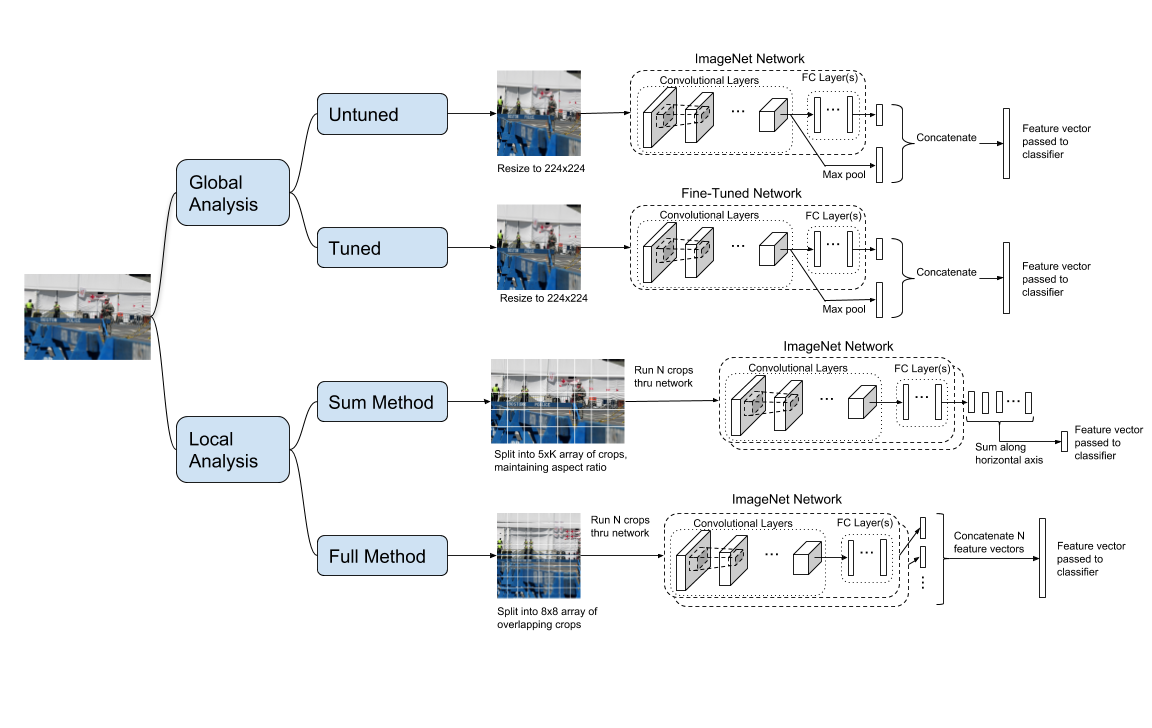}
\vspace{-15pt}
\begin{scriptsize}
\caption{Block schematic of proposed approach: the two broad methods are based on analyzing an image at the global and local level. Within each of these categories, we further analyzed two approaches, one that considers fine-tuning to a dataset, and the other that analyzes the effect of spatial context.}
\label{fig:summary-blk}
\end{scriptsize}
\end{figure*}

A common approach to distribute false information is to select an authentic image from some previously recorded event which convincingly supports their message, and re-brand it as a current story. 
For example, during the times of storms and hurricanes, images from previous hurricanes are usually re-purposed and uploaded in social media to create a scare. 
If we have a database of previously recorded events, then we will be able to verify those images that have been re-purposed from older events.
In this paper we investigate several methods to automatically identify images for event verification and image re-purposing detection. 
We explore a transfer learning approach~\cite{transfer} to event verification by using a network pre-trained on the ImageNet dataset~\cite{imagenet_cvpr09}, but instead of using the network for image classification we use the network for event verification .  After observing many images from different events, certain features may stand out to distinguish between two similar classes. Similar locations, architectures, or identifying symbols may be associated with each. For example, marathon race bibs are generally consistent for all participants in a single event, and different from other races.
In this paper, we outline two approaches of applying pre-trained ImageNet network for event verification, one at the global image level and other at the local image level. 
At the global level, we compare the effect of fine tuning a pre-trained network to a particular dataset to a method that is not tuned to any dataset.
At the local level, we explore the effects of spatial context at smaller scales using one method that does averaging and another method without averaging.
Figure.~\ref{fig:summary-blk} provides a summary of the methods.
Our experiments on several datasets show that both approaches are promising for event verification and image re-purposing detection.

\section{Related Work}
\label{sec:rel-work}

While there are several works on image classification, retrieval, event classification, and more, very few works address the problem of event verification. 
At the time of writing, we are aware of two  recent works that are related to this problem~\cite{jaiswal,Sabir}. 
The first considers pairs of images and captions\cite{jaiswal}, and then trains a network to decide how consistent the caption is with the associated image.
The second work~\cite{Sabir} altered GPS coordinates, captions, and the actual image pixels for image re-purposing detection. 
In contrast to these methods, our approach operates only at the image pixel level and does not need any metadata, which may not be always available at hand.

We approach the event verification problem as a transfer learning classification problem and distinguish between a finite set of events by using pre-trained networks.
A recent paper~\cite{transfer} systematically tested how well features from different architectures transfer to other classification tasks. 
Their tests showed that when comparing two different architectures, ImageNet performance was only weakly correlated with performance on another task if the network is fixed.
However, this effect diminished once the network was fine-tuned \cite{transfer}. 

Our methods also do not address the possibility of modification to image pixels. 
This is a well studied problem, with many viable solutions~\cite{forgery_1,forgery_2,forgery_4,forgery_5} 
Our models are created under the assumption that they will be used with one of these image manipulation detectors, and that all images given to our detector are unmodified.

\begin{figure*}[t]
  \centering
  \begin{subfigure}{.22\textwidth}
  \centering
  \includegraphics[width=0.95\linewidth]{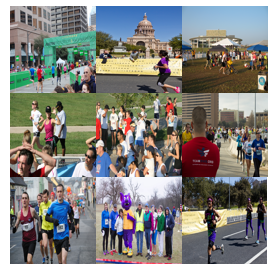}
  \caption{}
\end{subfigure}%
\begin{subfigure}{.22\textwidth}
  \centering
  \includegraphics[width=0.95\linewidth]{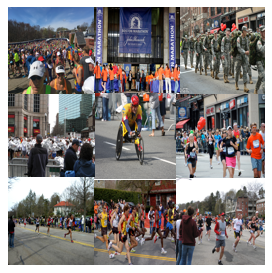}
  \caption{}
\end{subfigure}
\begin{subfigure}{.22\textwidth}
  \centering
  \includegraphics[width=0.95\linewidth]{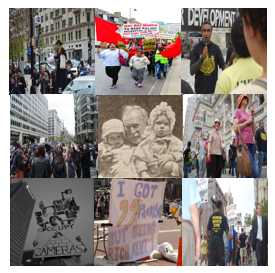}
  \caption{}
\end{subfigure}
\begin{subfigure}{.22\textwidth}
  \centering
  \includegraphics[width=0.95\linewidth]{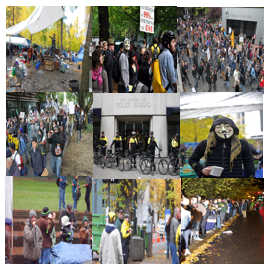}
  \caption{}
\end{subfigure}
\vspace{8pt}
\caption{Examples of images included in Ver$_1$ subset of the Event Verification dataset comprising the following events: (a) Austin Marathon, (b) Boston Marathon, (c) Occupy Baltimore, (d) Occupy Portland}
\label{fig:my_label}
\end{figure*}

\section{Event Verification and Image Repurposing Detection}
In this paper, our goal is to demonstrate an approach to verify if an image was taken at the claimed event. 
We explored two different approaches where one approach used global image features and another used local image features.
More specifically, we treated event verification as a transfer learning problem.
A deep learning classifier can be represented as a composition of the two function $f(\cdot) \in R^{N\times K}$ and $g(\cdot) \in R^{K \times M}$ where $f(\cdot)$ consists of the convolutional, non-linear, and pooling layers while $g(\cdot)$ is the fully connected classification layer with $M$ possible classes (see Figure \ref{fig:summary-blk}). 
During the training phase, the parameters of $f(\cdot)$ and $g(\cdot)$ were learned using a large image database such as ImageNet \cite{imagenet_cvpr09}.  

We investigated two transfer learning strategies that consisted of using the features from the pre-trained network and learning a new classifier using these features.  
The first strategy, or global method, extracted features by mapping the entire image using either $f(\cdot)$, $g(f(\cdot))$ or both.  
The second local method extracted features by mapping image patches using $g(f(\cdot))$, and the output from the patches were then averaged or concatenated into one large feature vector.  
A classifier was then trained on the extracted features to identify the respective events.
Finally, for the global method we also tested allowing the parameters of $f(\cdot)$ to be fine tuned using the event data training set. In total, there were four different strategies as illustrated in Figure $\ref{fig:summary-blk}$.    

\subsection{Global Method}
\noindent The global method consists of a two part structure. The first section uses deep convolutional neural networks (CNNs) to extract features from each image. 
The second is a separate classifier which takes these features as input.

\noindent \textbf{Untuned model}: The untuned network uses standard models such as ResNet-50 that are trained on ImageNet, and not tuned to the dataset under test. 
The images are first re-sized to the native resolution used by the CNN (either 224x224 or 299x299). 
The final feature vector is derived from the outputs from the last convolutional layer and the final output layer. 
This process replaces a large image with a much smaller feature vector, while hopefully retaining as much relevant information as possible. 
Using these features, we investigated several classifiers such as extra trees, random forests, nearest neighbors, support vector machines, and convolutional nets. 
The three variables tested were: the CNN used to extract features, which layers were used as features, and the final classifier algorithm.

\noindent \textbf{Tuned model}: We also investigated the effect of fine-tuning the CNN. A recent work suggested using at least 800 training images that contained at least 32 images per class~\cite{transfer}.
In our case we trained on 800 images with 200 per class, which is close to their minimum for total number of images.
The fine tuning was done in Keras, by removing the last fully connected layer of dimension 1000 used for ImageNet and replacing it with a fully connected layer with 4 outputs. 
An Adam optimizer with learning rate of 0.0001 was run and the training data was iterated through 10 times. 
Rates slower than 0.0001 did not show any improvement in validation accuracy while taking longer to converge. 
At our chosen rate, performance generally plateaued after 10 epochs.

\subsection{Local Method}
The local method differs from the global method in two key areas.  First, the local method uses a sliding window to extract features from local image patches.  Second, the local method derives its feature vector from the output of the fully connected classification layer. The goal with the local method is to capture more image details through the aggregation of information from local image patches. 

\noindent \textbf{Sum Features}: In order to deal with different image sizes, we rescaled each image to have 1120 rows while preserving the aspect ratio of the original image.  We then divided the image into overlapping 224x224 patches. Each block was processed through the entire ResNet-50 classifier \cite{he2016deep}. The output vectors for each patch were summed and the resulting sum was normalized to one, which produced a 1000 dimensional feature vector. This procedure extracted a feature vector of constant length from each image.  A classifier was then used to identify the event. Several classifiers were considered including support vector machines, extra trees, random forests, and xgboost~\cite{chen2016xgboost}.  

The sum feature vectors are similar to the bag-of-words image classification model~\cite{yang2007evaluating}.  
In particular, the output of each patch can be interpreted as a distribution over classes, which is a 1000 dimensional vector of words or phrases.  The words for each patch are then summed, generating a bag of words.  The original bag-of-words model would identify one word per patch, however our approach outputs a probability distribution of words for each patch and we sum the different probability distribution to generate our feature vector.    


\noindent \textbf{Full Features}: The sum features average the information gained from all the patches and removes any spatial context.  As a simple test to investigate the importance of the spatial knowledge, we removed the sum over the patches. In order to obtain feature vectors that had consistent dimensions, we rescaled each image to have 1120 rows and 1120 columns.  ResNet-50 and a sliding window was used to produce a feature vector for each 224x224 block with an overlap of 100 pixels.  The final output was the concatenation of the output of all the feature vectors.  The same classifiers were tested as in the sum features.

\section{Experiments and Results}
Here we will detail the datasets used in our experiments, the results on the global and local methods and a comparison of both the methods. 
For each method, a receiver operating curve (ROC) was obtained, and the area under this curve (AUC) is used as the primary metric for comparison.

\subsection{Event Verification (EV) Dataset}
The classification procedures were tested on the Event Verification dataset generated by NIST as part of the DARPA Media Forensics (MediFor) project.
This dataset contained three different subsets, each already divided into training and testing. 
We refer to them as Ver$_1$, Ver$_2$ and Ver$_{eval}$. 
The first two versions each had 4 events, with 200 training images for each event, and 100 testing images per event. 
Ver$_1$ subset contained images associated with the Boston Marathon, Austin Marathon, Occupy Portland, and Occupy Baltimore (see Figure~\ref{fig:my_label}). 
Ver$_2$ had images from Hurricane Sandy, Hurricane Matthew, the Oshkosh Air Show, and Berlin Air Show.
The third set was held out for evaluation (Ver$_{eval}$). 
This subset had 12 events, with 200 images per event, and 600 total test images with held out labels. 
Testing on Ver$_{eval}$ subset was done by submission of results to a NIST server, which in turn returned the ROC curve. 
Ver$_1$, Ver$_2$ subsets were used to test across a wide number of models, while Ver$_{eval}$ subset was reserved to see how well these models generalized.

\begin{figure*}[h]
  \centering
  \begin{subfigure}{.3\textwidth}
  \centering
  \includegraphics[width=0.8\linewidth, height=0.6\columnwidth]{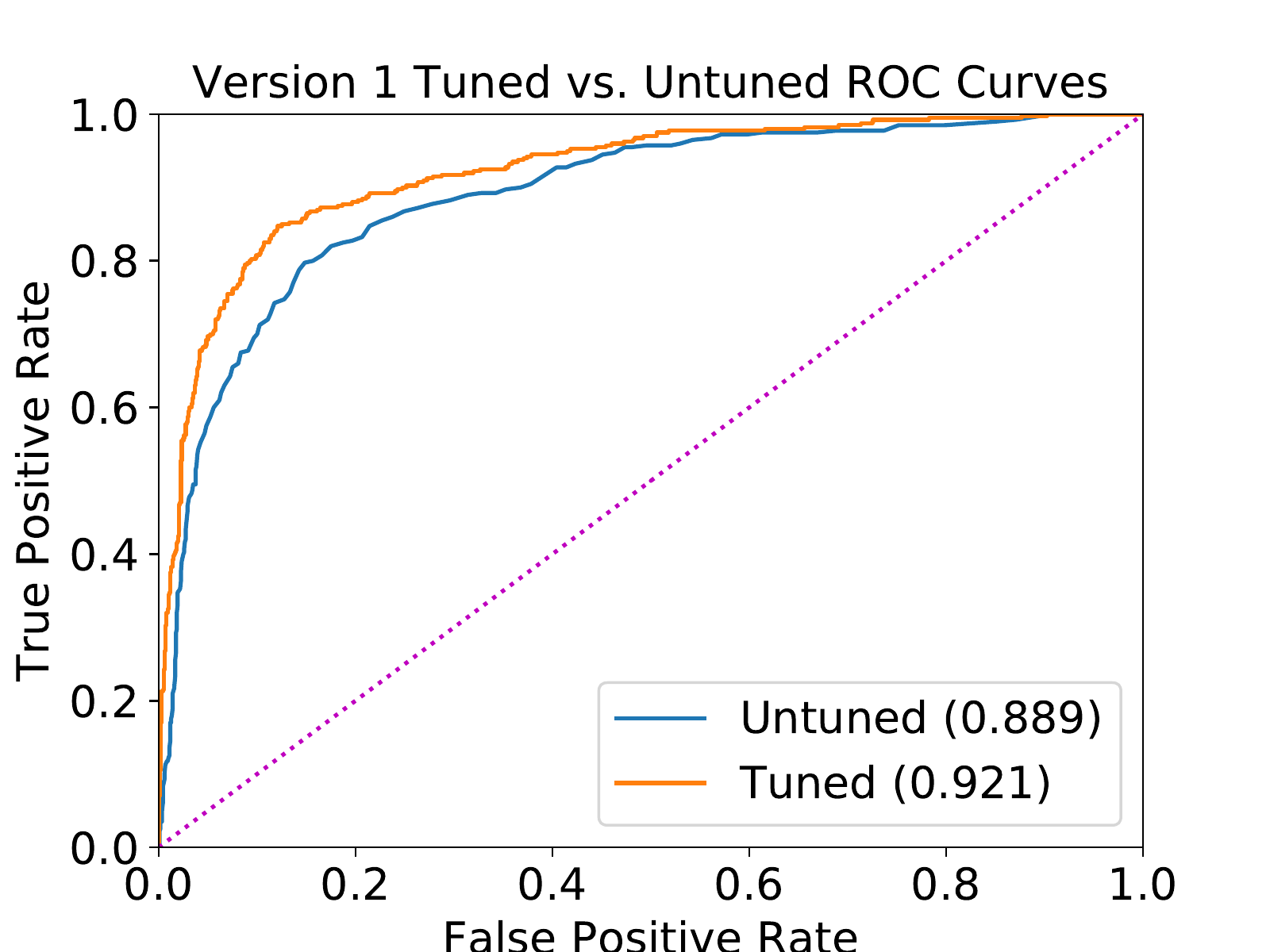}
\end{subfigure}%
\begin{subfigure}{.3\textwidth}
  \centering
  \includegraphics[width=0.8\linewidth, height=0.6\columnwidth]{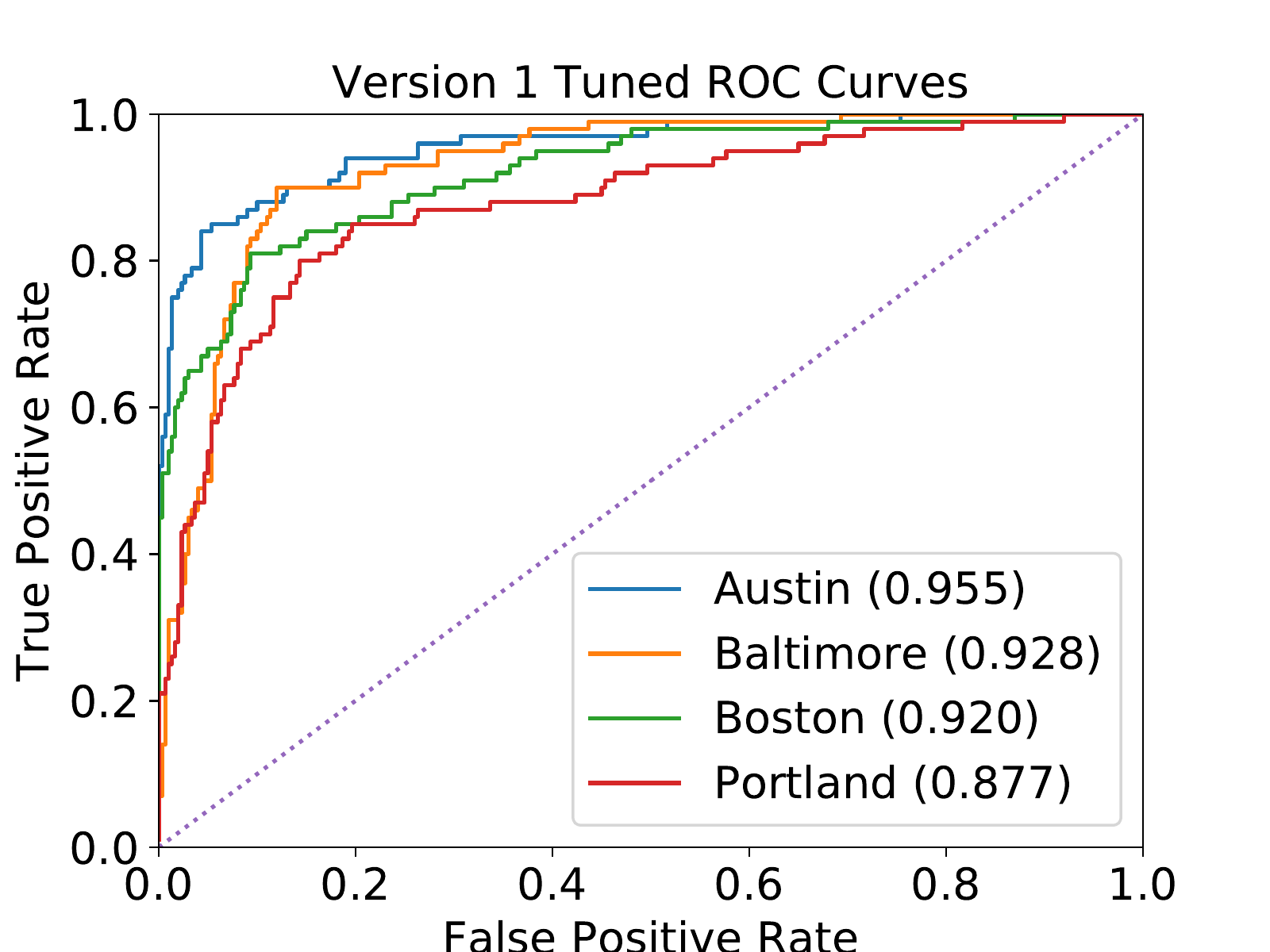}
\end{subfigure}
\begin{subfigure}{.3\textwidth}
  \centering
  \includegraphics[width=0.8\linewidth, height=0.6\columnwidth]{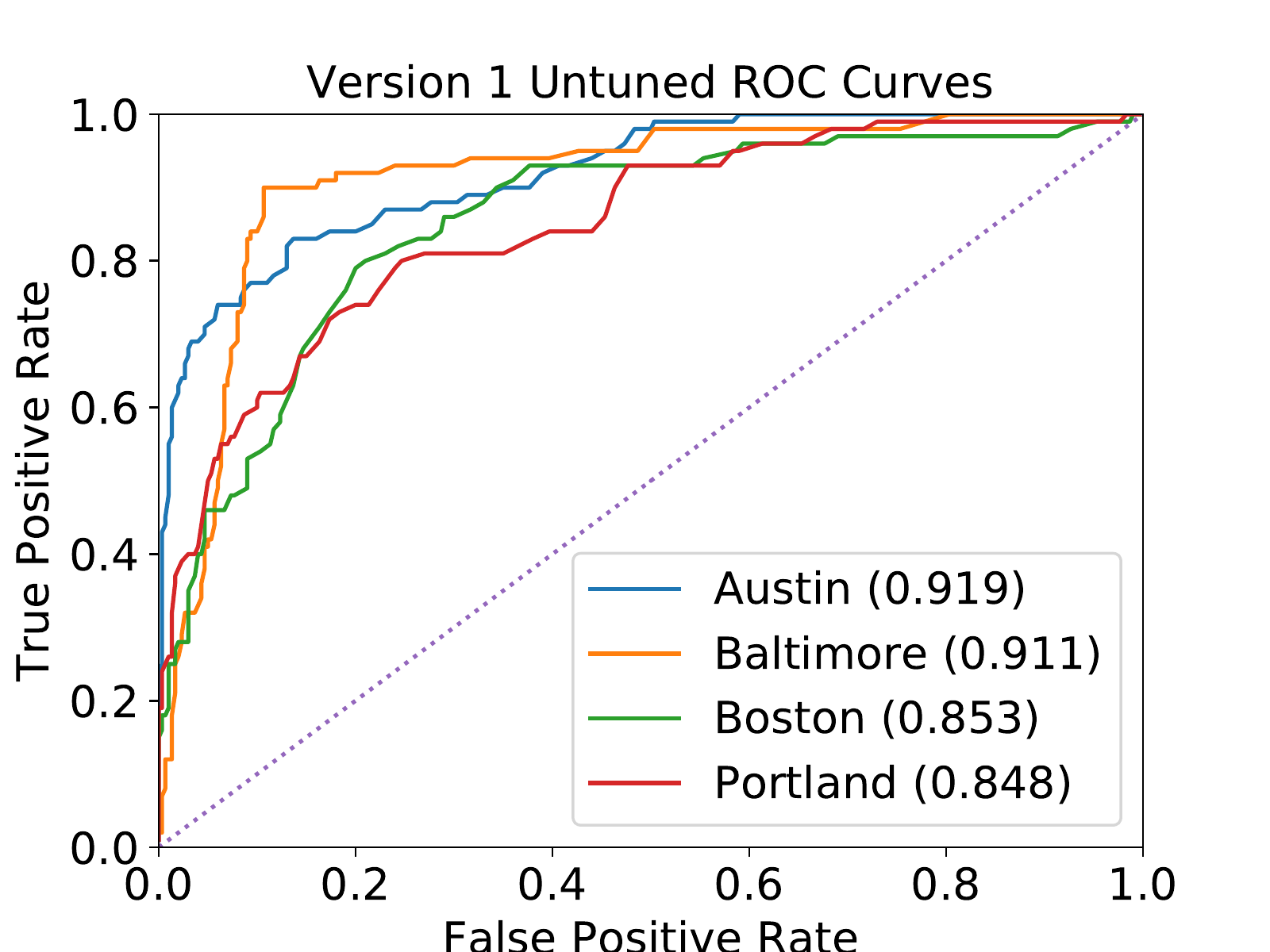}
\end{subfigure}
\vspace{3pt}
\caption{ROC curves using the global features and tested on the Ver$_{1}$ subset.}
\label{fig:ver1-global}
\end{figure*}

\begin{figure*}[h]
  \centering
  \begin{subfigure}{.3\textwidth}
  \centering
  \includegraphics[width=0.8\linewidth, height=0.6\columnwidth]{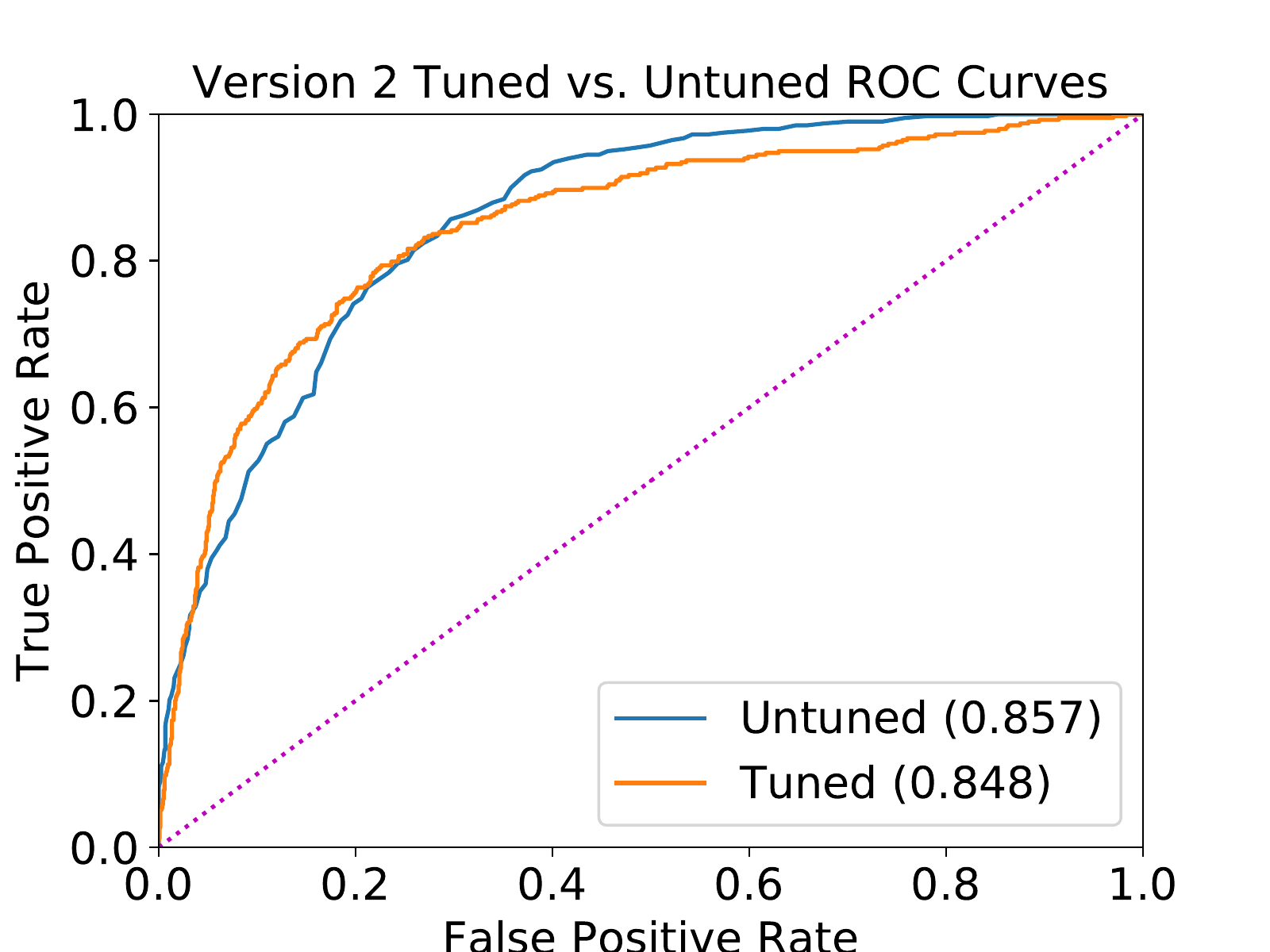}
\end{subfigure}%
\begin{subfigure}{.3\textwidth}
  \centering
  \includegraphics[width=0.8\linewidth, height=0.6\columnwidth]{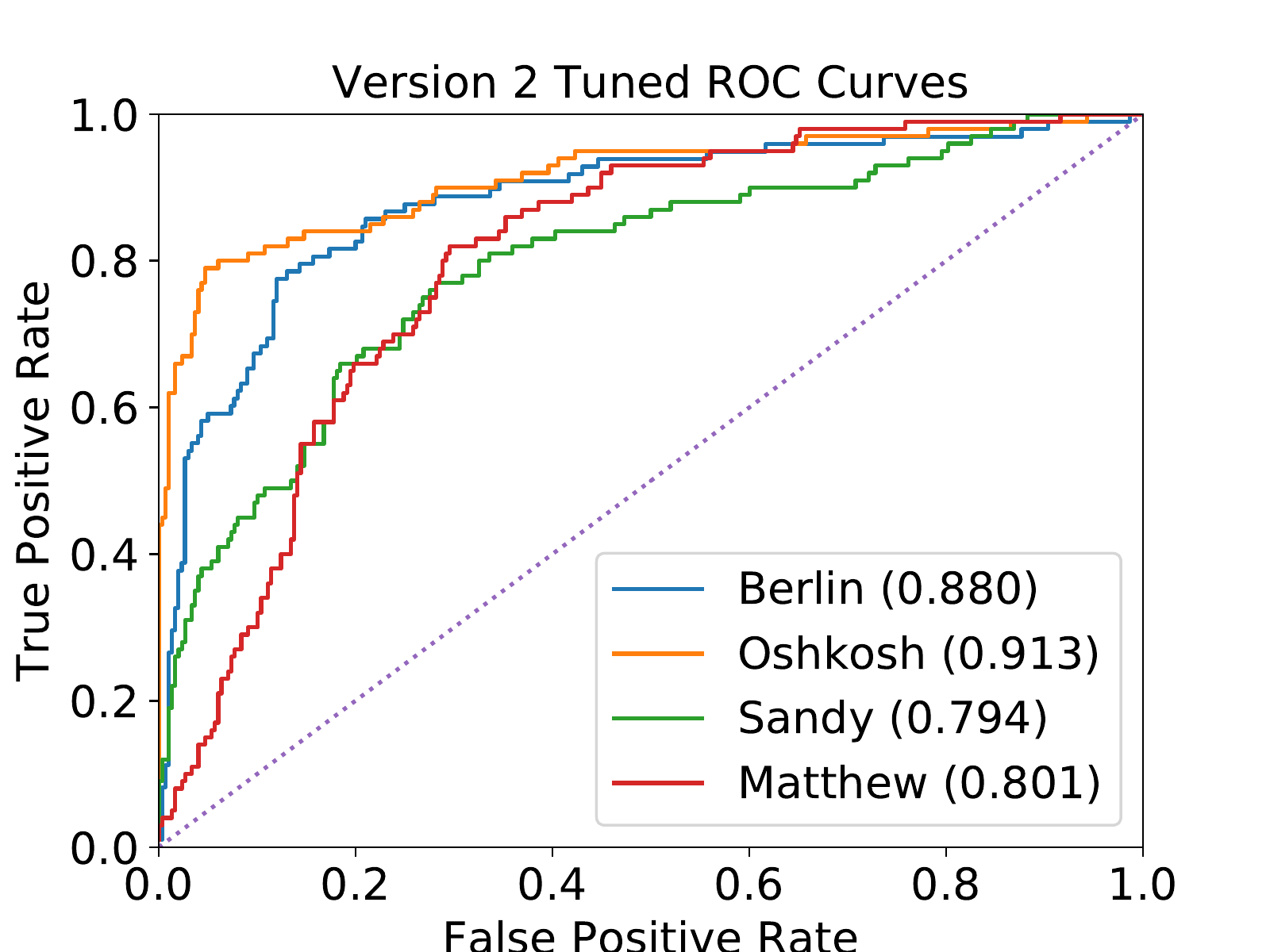}
\end{subfigure}
\begin{subfigure}{.3\textwidth}
  \centering
  \includegraphics[width=0.8\linewidth, height=0.6\columnwidth]{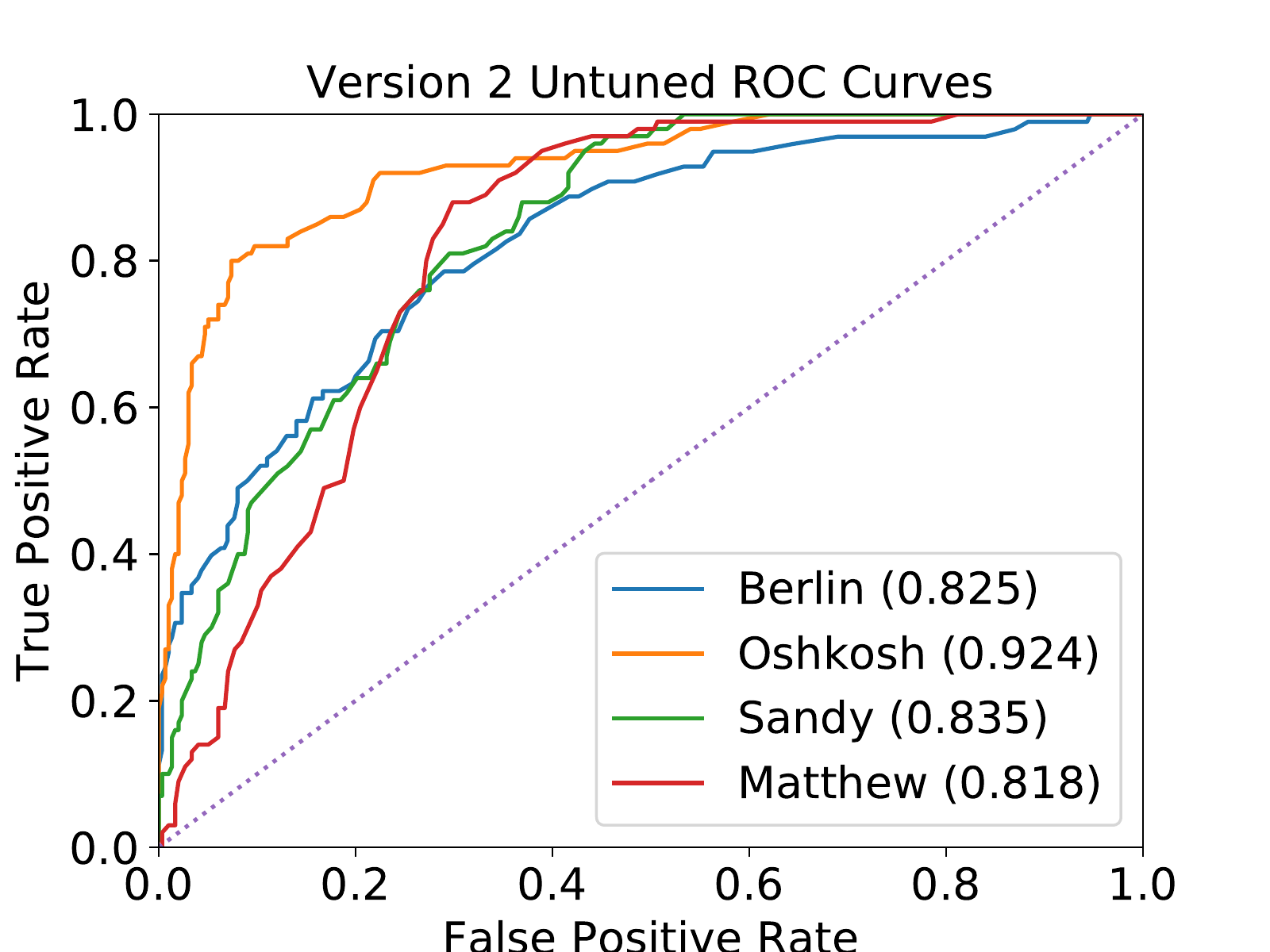}
\end{subfigure}
\vspace{3pt}
\caption{ROC curves using the global features and tested on the Ver$_{2}$ subset.}
\label{fig:ver2-global}
\end{figure*}

\subsection{Results on Global Method}

For the global method, many combinations of ImageNet CNNs, feature extraction locations, and standard machine learning classifiers were tested.
In general, using only intermediate layer features performed better than only considering output layer features.
Using both gave a slight boost for some classifiers though the difference was insignificant for the best classifiers. 
The results presented for the rest of this section are from the intermediate layer. 
For most cases, ResNet-50 had the best performance, which is consistent with the results from the recent work on transfer learning~\cite{transfer}.
For a network strictly trained on ImageNet, the classifier at the end had a strong impact on performance. 
Random Forest and Extra Trees classifiers performed the best in most cases. On Ver$_1$, the fine-tuned ResNet universally outperformed the top untuned method. 
For Ver$_2$ the results were less clear as AUC scores were close. 
At low false alarm rates the tuned method performs better, while the untuned method has better detection at high false alarm rates. 
AUC is slightly higher for the untuned case overall and per class ROC curves are shown in Figure~\ref{fig:ver1-global} for Ver$_1$ subset and Figure~\ref{fig:ver2-global} for Ver$_2$ subset.

After fine tuning, the classifier at the end seemed to have less of an impact. 
The best performing of the fine-tuned classifiers were simply the dense output layer generated during the tuning process. 
Using a different classifier in the end may not have much benefit.
However, distance based classifiers do have the ability to provide training images which it found to be similar to the query image, and can be helpful in some applications where justification is needed.
More detailed experiments on the effect of different classifiers on various pre-trained networks are presented later in Table.~\ref{tab:global_results_array}.

\begin{table}[b]
\centering
\scriptsize
\caption{Local method AUC results for different classifiers and different features on the Ver$_{1}$ subset.  From left to right the classifiers are extra trees, random forest, 1 nearest neighbor with Euclidean distance, support vector machines, and xgboost.}
\vspace{2pt}
\begin{tabular}{|c|c|c|c|c|c|}
\hline
Features & ET & RF & 1-NN & SVM & XGB\\
\hline
Local Sum & \textbf{0.885}\cellcolor{black!46.0} & .882\cellcolor{black!45.0} & 0.74\cellcolor{black!40.0} & 0.855\cellcolor{black!42.0} & .872\cellcolor{black!44.0} \\
\hline
Local Full & \textbf{0.857}\cellcolor{black!42.0} & 0.851\cellcolor{black!42.0} & 0.612\cellcolor{black!13.0} & 0.852\cellcolor{black!42.0} & .841\cellcolor{black!41.0}\\
\hline
\end{tabular}
\label{tab:local}
\end{table}

\begin{table}[b]
\centering
\scriptsize
\caption{Comparison of AUC between Global and Local Method}
\vspace{2pt}
\begin{tabular}{|c|c|c|c|c|}
\hline
Dataset & Global Untuned & Global Tuned & Local Sum & Local Full \\
\hline
Ver$_1$ & 0.889\cellcolor{black!46.0} & \textbf{0.921}\cellcolor{black!50.0} & 0.885\cellcolor{black!46.0} & 0.857\cellcolor{black!42.0}\\
\hline
Ver$_2$ & \textbf{0.857}\cellcolor{black!42.0} & 0.848\cellcolor{black!41.0} & 0.831\cellcolor{black!39.0} & 0.798\cellcolor{black!35.0}\\
\hline
Ver$_{eval}$ & 0.88\cellcolor{black!45.0} & \textbf{0.89}\cellcolor{black!46.0} & 0.85\cellcolor{black!42.0} & 0.82\cellcolor{black!38.0}\\
\hline
\end{tabular}
\label{tab:comparison}
\end{table}


\begin{figure*}[t]
  \centering
  \begin{subfigure}{.3\textwidth}
  \centering
  \includegraphics[width=0.8\linewidth, height=0.6\columnwidth]{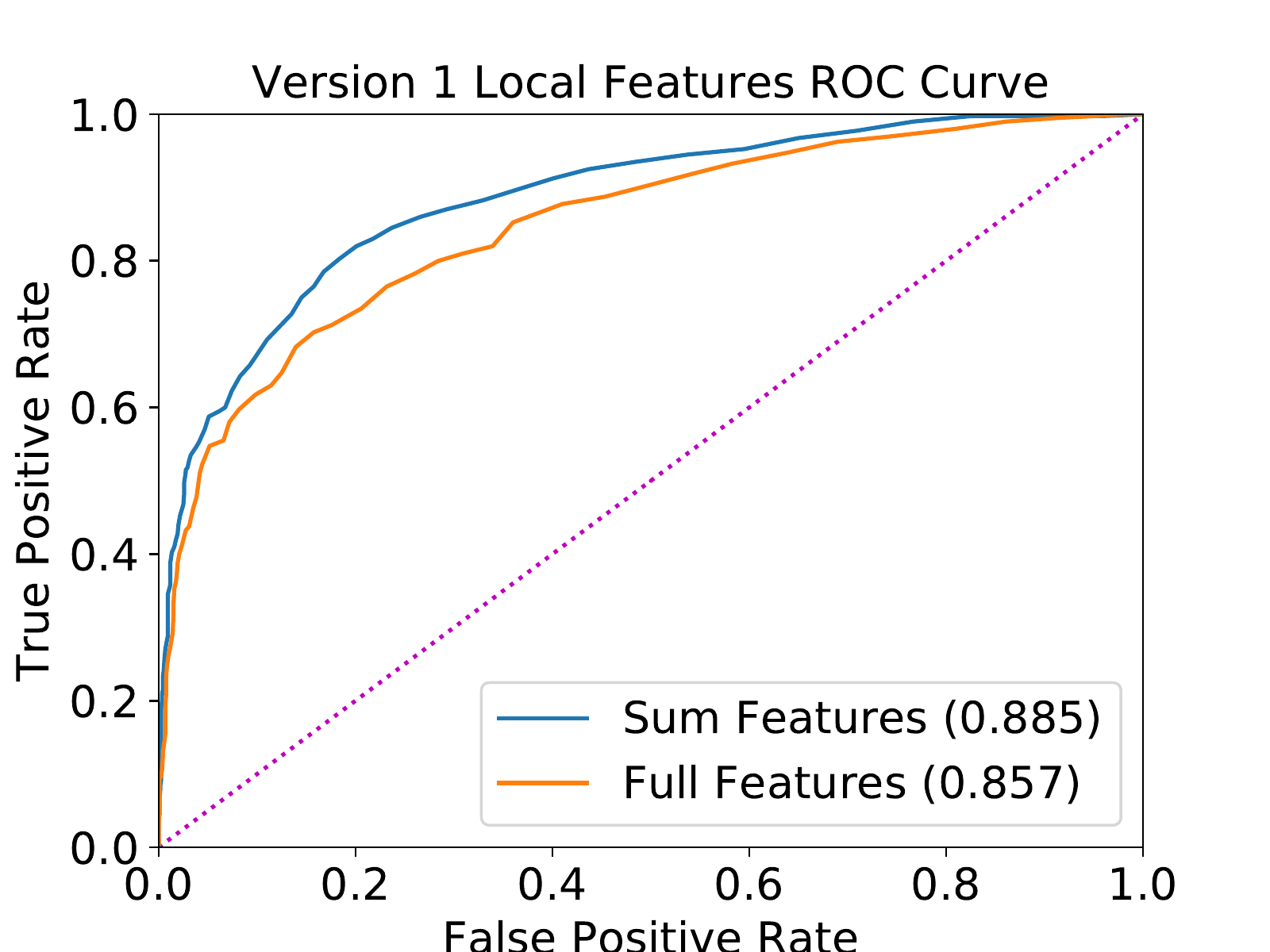}
\end{subfigure}%
\begin{subfigure}{.3\textwidth}
  \centering
  \includegraphics[width=0.8\linewidth, height=0.6\columnwidth]{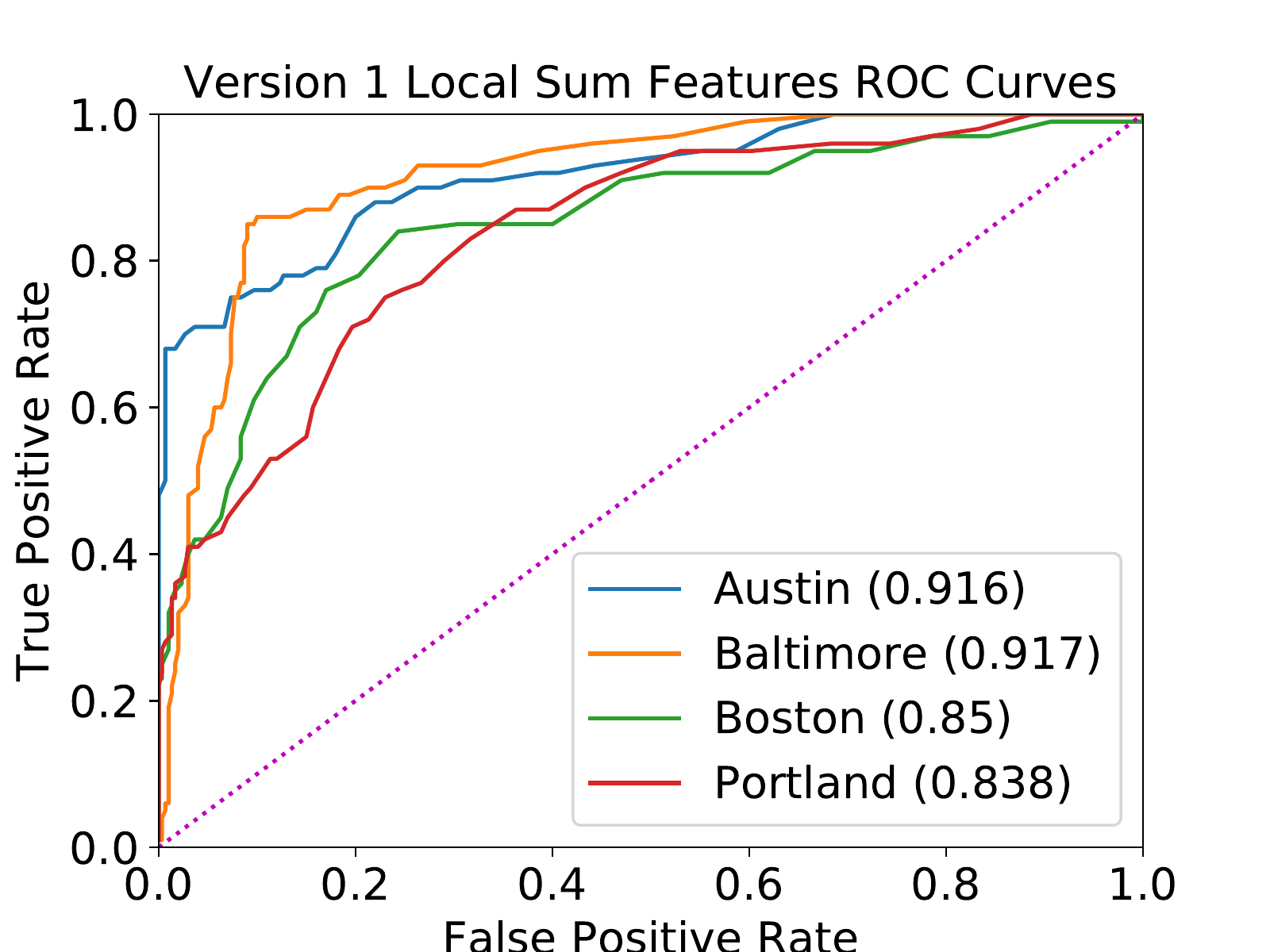}
\end{subfigure}
\begin{subfigure}{.3\textwidth}
 \centering
 \includegraphics[width=0.8\linewidth, height=0.6\columnwidth]{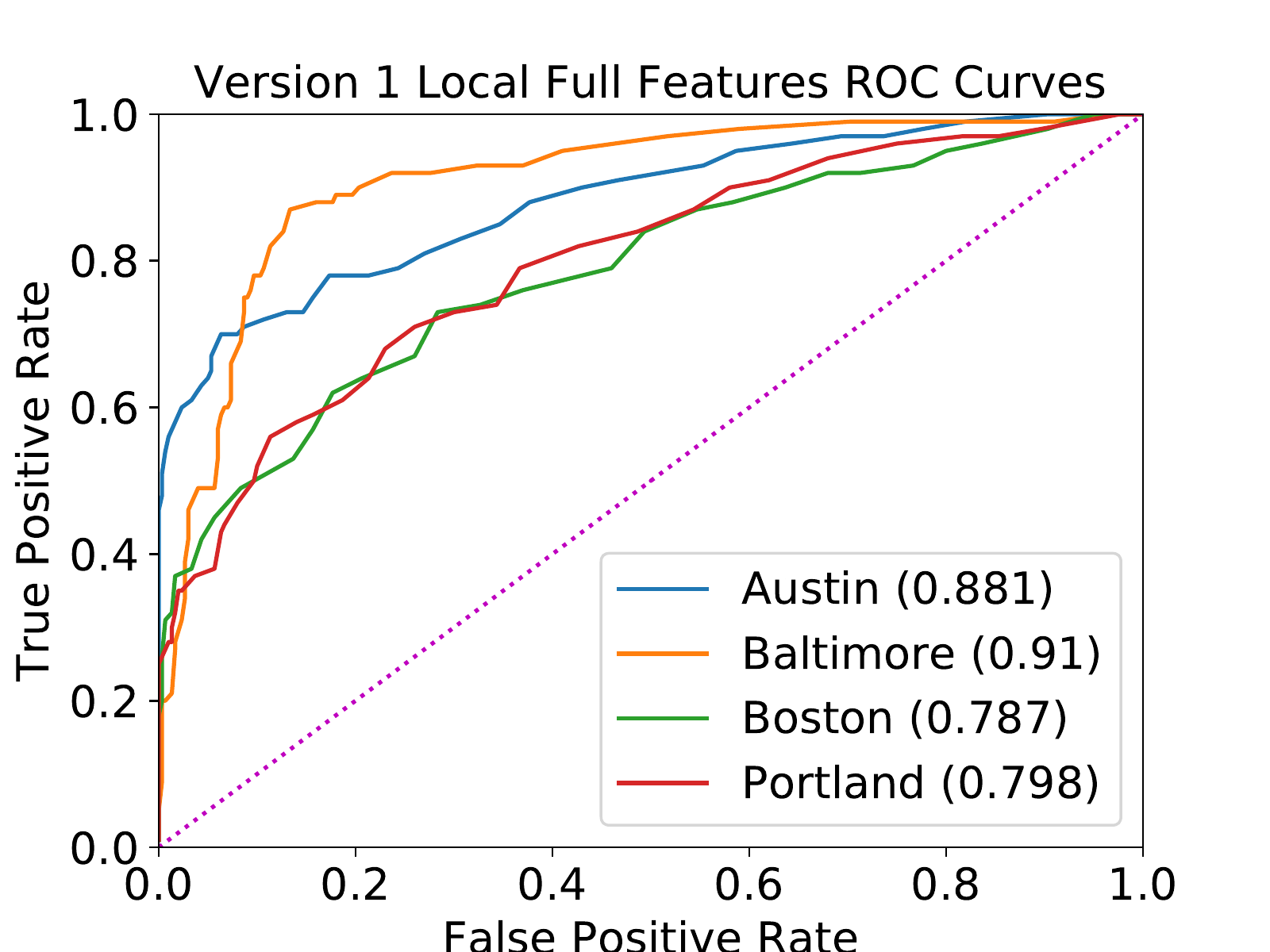}
\end{subfigure}
\vspace{3pt}
\caption{ROC curves using the local features and tested on Ver$_{1}$ subset.}
\label{fig:ver1-local}
\end{figure*}

\begin{figure*}[t]
  \centering
  \begin{subfigure}{.3\textwidth}
  \centering
  \includegraphics[width=0.8\linewidth, height=0.6\columnwidth]{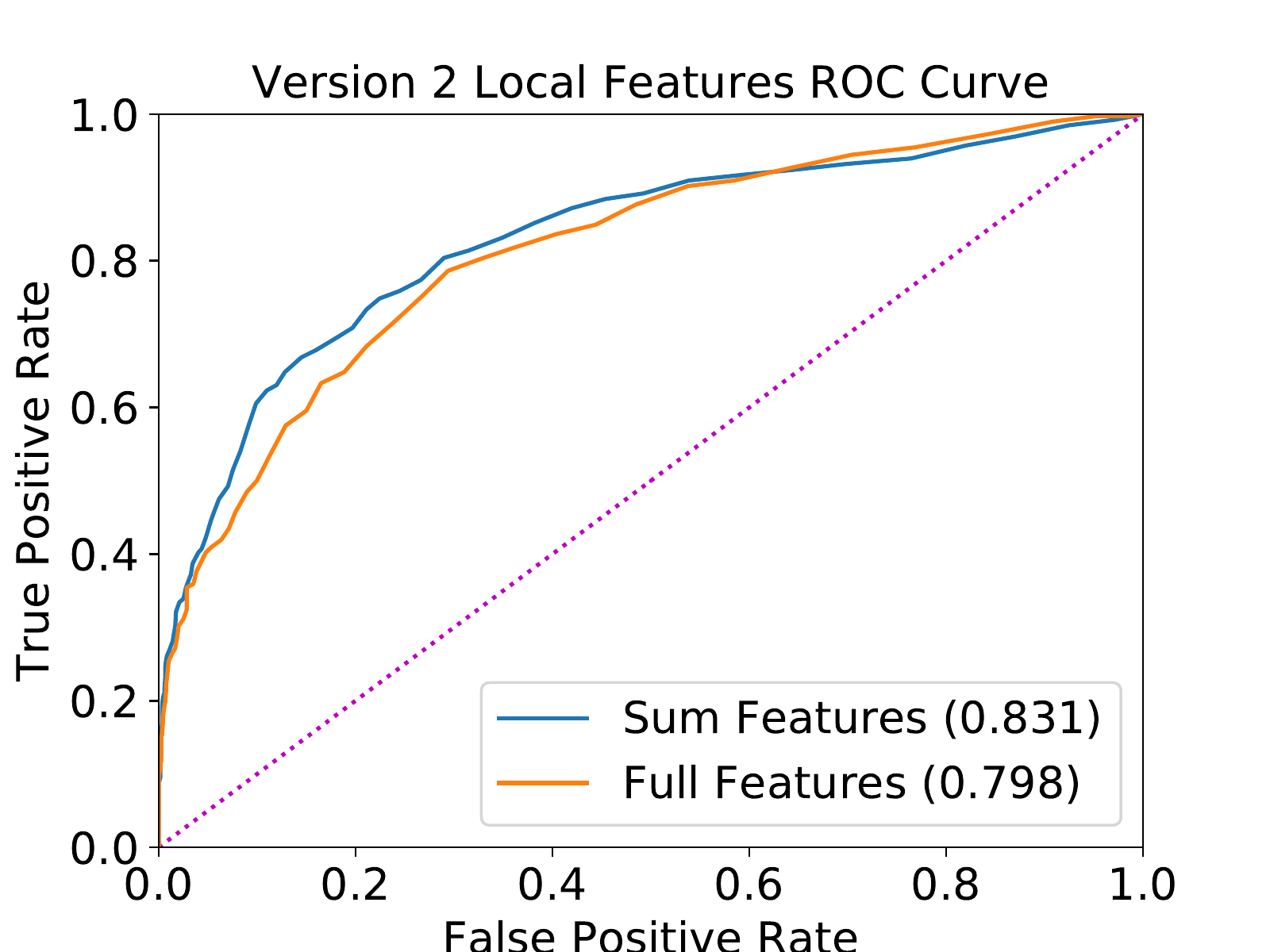}
\end{subfigure}%
\begin{subfigure}{.3\textwidth}
  \centering
  \includegraphics[width=0.8\linewidth, height=0.6\columnwidth]{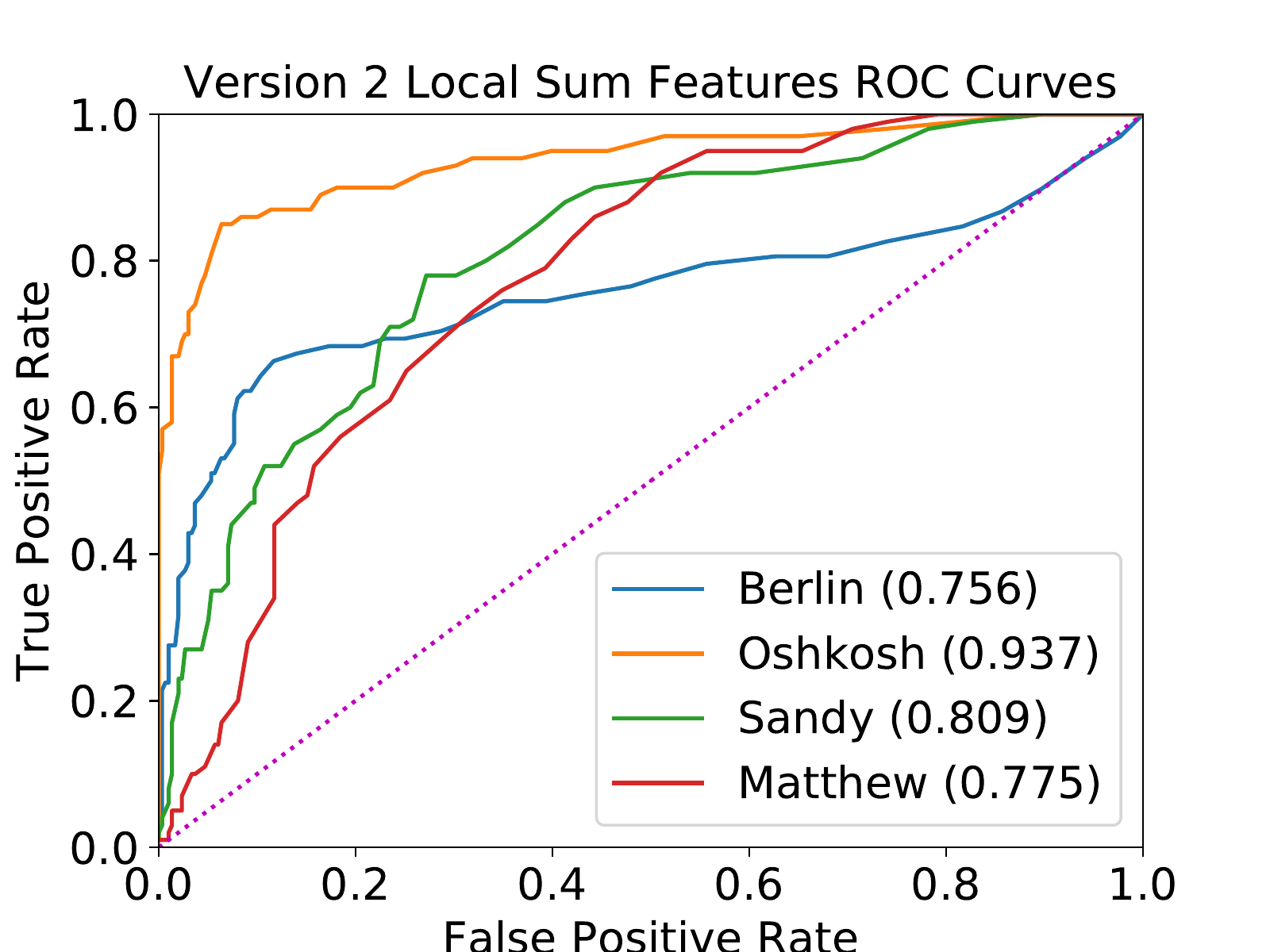}
\end{subfigure}
\begin{subfigure}{.3\textwidth}
 \centering
 \includegraphics[width=0.8\linewidth, height=0.6\columnwidth]{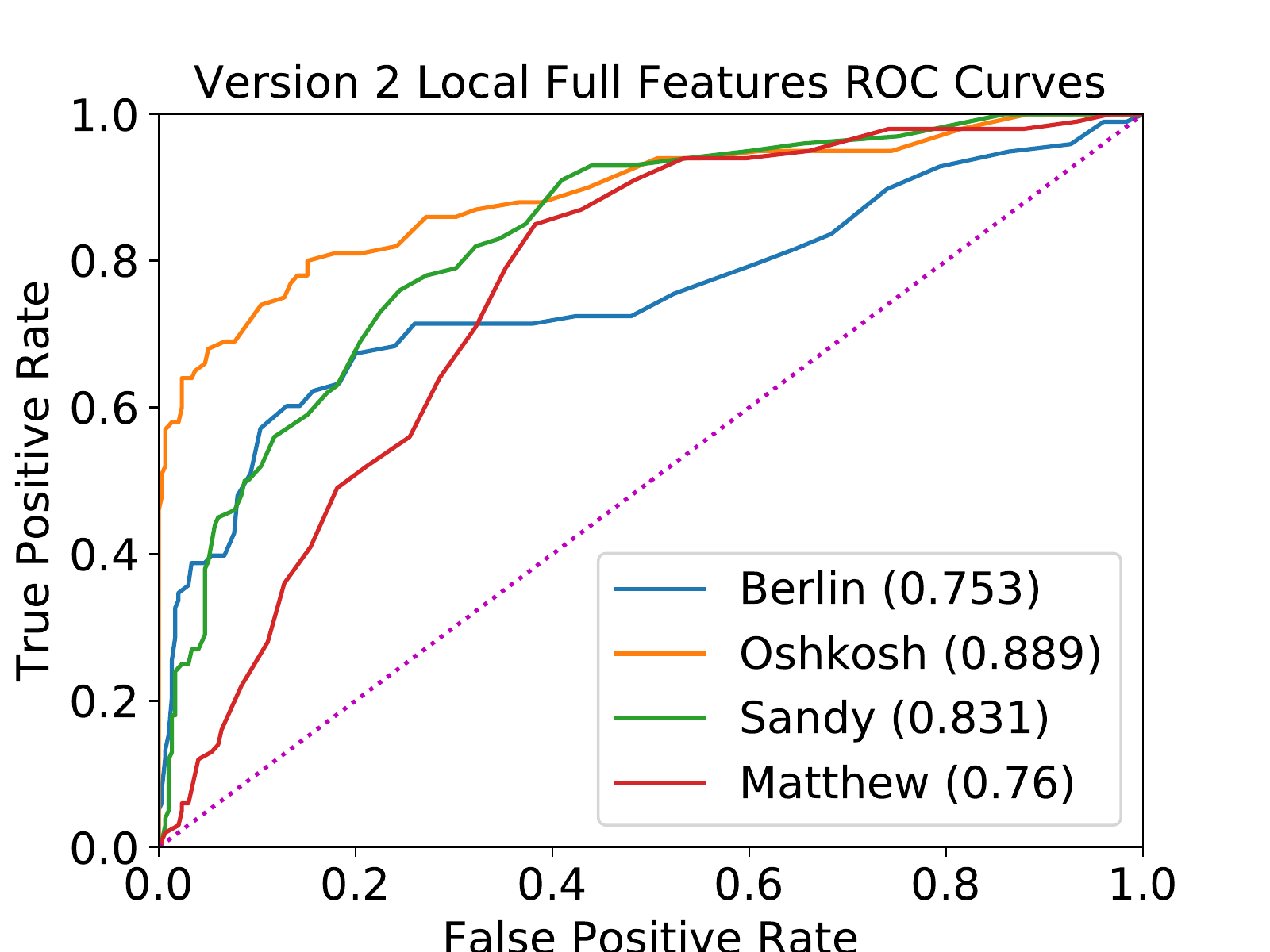}
\end{subfigure}
\vspace{3pt}
\caption{ROC curves using the local features and tested on the Ver$_{2}$ subset.}
\label{fig:ver2-local}
\end{figure*}

\subsection{Results on Local Method}

Based on the results of the global method, only the ResNet-50 network trained using ImageNet was investigated.
The classifiers tested were extra trees, random forests, support vector machines, one nearest neighbor, and xgboost which are summarized for the Ver$_{1}$ subset in Table~\ref{tab:local}. 
The ensemble methods of extra trees and random forest performed the best for the sum features and were statistically equivalent.  
For the full features, the extra trees, random forest, and support vector machines had equivalent results.
The extra trees and random forest ensemble classifiers performed better on the sum features than the full features, while the support vector machine results was equivalent for the sum and full features. 
As with the global method, due to random initialization, the results using the same classifier would fluctuate by 1\%. 
The local method was tested on the Ver$_1$, Ver$_2$ and Ver$_{eval}$ subsets of the Event Verification dataset. 
For all the classifiers except xgboost, the python package scikit-learn version 0.19.1 was used and the default settings obtained the best performance. 
The python package xgboost was used for the xgboost implementation with a max depth of two and binary logistic objective. 
The ROC curves for this method are shown in Figures~\ref{fig:ver1-local} and~\ref{fig:ver2-local} for the Ver$_{1}$ and Ver$_{2}$ subsets.

\subsection{Comparison of Global and Local Method}

\begin{figure}[t]
\centering
\includegraphics[width=.90\linewidth]{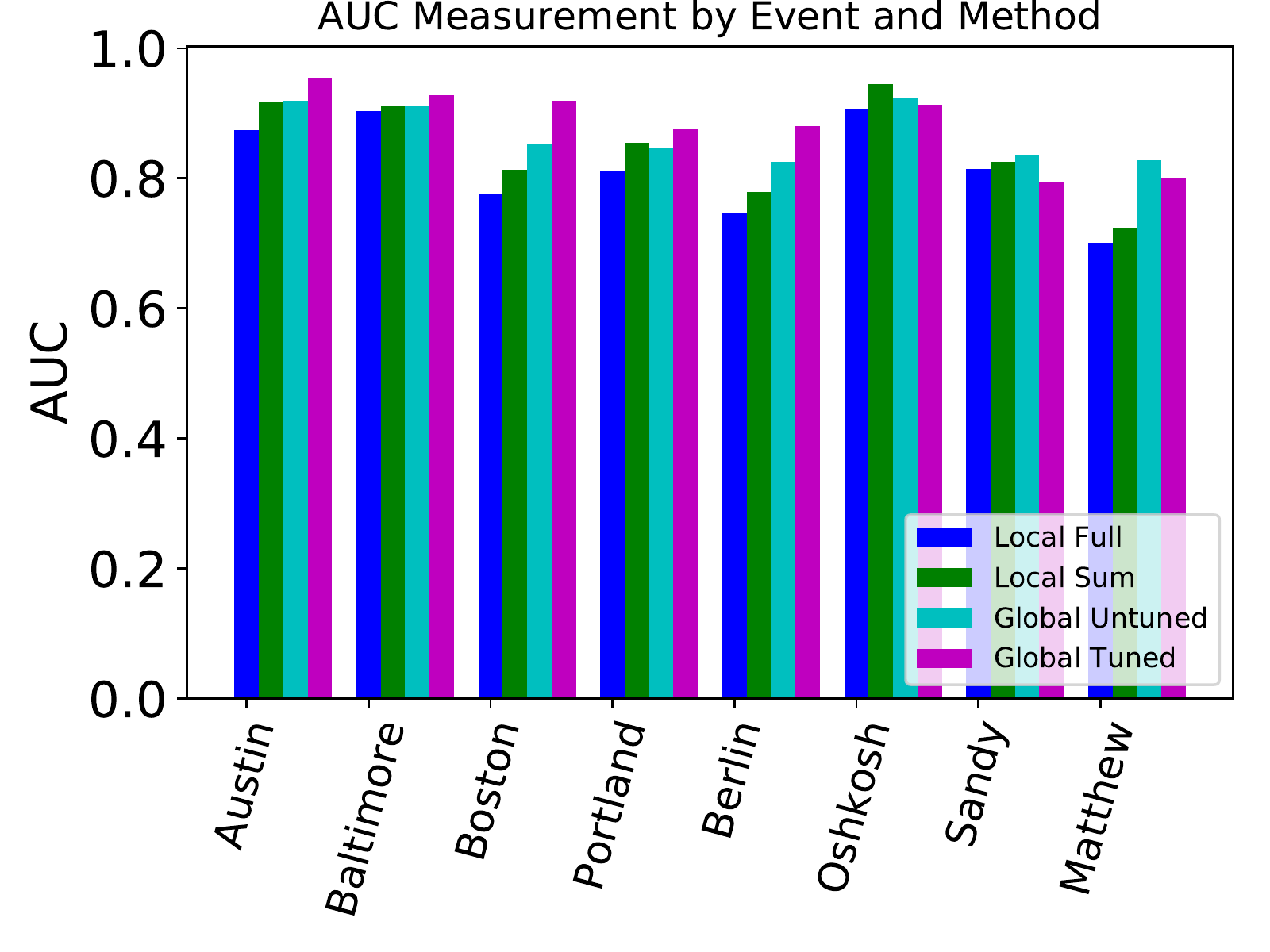}
\caption{Comparison of AUCs on individual events on Ver$_1$ and Ver$_2$ subsets of the Event Verification Dataset.}
\label{fig:auc-comp}
\end{figure}

The comparison of the global method and the local method on Ver$_1$, Ver$_2$ and Ver$_{eval}$ subsets of the Event Verification dataset are summarized in Table~\ref{tab:comparison} and Figure~\ref{fig:auc-comp}.
On average, we can see that the global methods perform better than the local methods.
Among the global methods, the tuned model obtained the highest AUC for Ver$_1$ and Ver$_{eval}$ subsets while the untuned model performed slightly better than the tuned model for the Ver$_2$ subset. 
In general, we see that the tuned model would be the most preferable though it would come at the extra cost of tuning the model for every dataset. 
Next we analyzed the performance of individual events for the four methods. The results are summarized for both the Ver$_1$ and Ver$_2$ subsets in Figure.~\ref{fig:auc-comp}. 
The global tuned model performed the best for most events (Austin Marathon, Occupy Baltimore, Boston Marathon, Occupy Portland and Berlin Airshow), while the global untuned model performed the best for two events (Hurricane Sandy, Hurricane Matthew). 
While the global method tends to outperform the local method in most events, there are cases in which one may be preferred. In the one event where the local method outperformed the global one, Oshkosh Airshow, we hypothesize that there was more detail at the smaller scale. 
Also, for events such as Occupy Baltimore and Hurricane Sandy, there isn't any significant difference in the performance of the four methods. 
Another consideration is that all of the methods besides local sum resize the image to be square. If preserving the aspect ratio is important to a particular task, this may perform relatively better. Testing on several different datasets also showed that performance is highly data dependant. Given our results, we expect that the best out-of-the-box approach is either the global tuned or global untuned.

\begin{table*}[ht]
\caption{AUC results for various combinations of CNN, feature extraction locations, and end classifiers. The classifiers on the top row are: Extra Trees, Random Forest, 1 Nearest Neighbor (NN) with L1 distance, 1 NN with L2 distance, 2 NN with L2, 4 NN with L2.}
\label{tab:global_results_array}
\vspace{0.5cm}
\begin{scriptsize}
\begin{minipage}[b]{0.48\linewidth}
\centering
\caption*{Features only from intermediate layer}
\begin{tabular}{|c||c|c|c|c|c|c|}
\hline
 & ET & RF & L1 & L2 & 2NN & 4NN\\
\hline
\hline
Xception & 0.766\cellcolor{black!32.0} & 0.753\cellcolor{black!30.0} & 0.641\cellcolor{black!17.0} & 0.638\cellcolor{black!16.0} & 0.675\cellcolor{black!21.0} & 0.695\cellcolor{black!23.0} \\
\hline
VGG16 & 0.880\cellcolor{black!45.0} & 0.874\cellcolor{black!44.0} & 0.751\cellcolor{black!30.0} & 0.775\cellcolor{black!33.0} & 0.818\cellcolor{black!38.0} & 0.845\cellcolor{black!41.0} \\
\hline
VGG19 & 0.883\cellcolor{black!46.0} & 0.874\cellcolor{black!44.0} & 0.756\cellcolor{black!30.0} & 0.768\cellcolor{black!32.0} & 0.820\cellcolor{black!38.0} & 0.838\cellcolor{black!40.0} \\
\hline
ResNet50 & 0.889\cellcolor{black!46.0} & 0.886\cellcolor{black!46.0} & 0.79\cellcolor{black!34.0} & 0.781\cellcolor{black!33.0} & 0.839\cellcolor{black!40.0} & 0.877\cellcolor{black!45.0} \\
\hline
InceptionV3 & 0.765\cellcolor{black!31.0} & 0.772\cellcolor{black!32.0} & 0.648\cellcolor{black!17.0} & 0.655\cellcolor{black!18.0} & 0.702\cellcolor{black!24.0} & 0.714\cellcolor{black!25.0} \\
\hline
InceptionResNet & 0.682\cellcolor{black!21.0} & 0.695\cellcolor{black!23.0} & 0.575\cellcolor{black!9.0} & 0.573\cellcolor{black!8.0} & 0.586\cellcolor{black!10.0} & 0.614\cellcolor{black!13.0} \\
\hline
MobileNet & 0.842\cellcolor{black!41.0} & 0.847\cellcolor{black!41.0} & 0.739\cellcolor{black!28.0} & 0.735\cellcolor{black!28.0} & 0.786\cellcolor{black!34.0} & 0.825\cellcolor{black!39.0} \\
\hline
MobileNet2 & 0.887\cellcolor{black!46.0} & 0.876\cellcolor{black!45.0} & 0.766\cellcolor{black!32.0} & 0.753\cellcolor{black!30.0} & 0.805\cellcolor{black!36.0} & 0.843\cellcolor{black!41.0} \\
\hline
DenseNet121 & 0.877\cellcolor{black!45.0} & 0.871\cellcolor{black!44.0} & 0.745\cellcolor{black!29.0} & 0.73\cellcolor{black!27.0} & 0.785\cellcolor{black!34.0} & 0.819\cellcolor{black!38.0} \\
\hline
DenseNet169 & 0.869\cellcolor{black!44.0} & 0.870\cellcolor{black!44.0} & 0.748\cellcolor{black!29.0} & 0.745\cellcolor{black!29.0} & 0.787\cellcolor{black!34.0} & 0.810\cellcolor{black!37.0} \\
\hline
DenseNet201 & 0.735\cellcolor{black!28.0} & 0.728\cellcolor{black!27.0} & 0.636\cellcolor{black!16.0} & 0.636\cellcolor{black!16.0} & 0.655\cellcolor{black!18.0} & 0.690\cellcolor{black!22.0} \\
\hline
NASNetMobile & 0.764\cellcolor{black!31.0} & 0.763\cellcolor{black!31.0} & 0.648\cellcolor{black!17.0} & 0.646\cellcolor{black!17.0} & 0.680\cellcolor{black!21.0} & 0.701\cellcolor{black!24.0} \\
\hline
Tuned ResNet & 0.909\cellcolor{black!49.0} & 0.905\cellcolor{black!48.0} & 0.845\cellcolor{black!41.0} & 0.831\cellcolor{black!39.0} & 0.867\cellcolor{black!44.0} & 0.876\cellcolor{black!45.0} \\
\hline
\end{tabular}
\end{minipage}
\hspace{0.5cm}
\begin{minipage}[b]{0.48\linewidth}
\centering
\caption*{Features from intermediate and output layers}
\begin{tabular}{|c||c|c|c|c|c|c|}
\hline
 & ET & RF & L1 & L2 & 2NN & 4NN\\
\hline
\hline
Xception & 0.760\cellcolor{black!31.0} & 0.756\cellcolor{black!30.0} & 0.641\cellcolor{black!17.0} & 0.638\cellcolor{black!16.0} & 0.675\cellcolor{black!21.0} & 0.695\cellcolor{black!23.0} \\
\hline
VGG16 & 0.879\cellcolor{black!45.0} & 0.876\cellcolor{black!45.0} & 0.751\cellcolor{black!30.0} & 0.775\cellcolor{black!33.0} & 0.818\cellcolor{black!38.0} & 0.845\cellcolor{black!41.0} \\
\hline
VGG19 & 0.884\cellcolor{black!46.0} & 0.884\cellcolor{black!46.0} & 0.756\cellcolor{black!30.0} & 0.768\cellcolor{black!32.0} & 0.820\cellcolor{black!38.0} & 0.838\cellcolor{black!40.0} \\
\hline
ResNet50 & 0.894\cellcolor{black!47.0} & 0.895\cellcolor{black!47.0} & 0.79\cellcolor{black!34.0} & 0.781\cellcolor{black!33.0} & 0.839\cellcolor{black!40.0} & 0.877\cellcolor{black!45.0} \\
\hline
InceptionV3 & 0.762\cellcolor{black!31.0} & 0.769\cellcolor{black!32.0} & 0.648\cellcolor{black!17.0} & 0.655\cellcolor{black!18.0} & 0.702\cellcolor{black!24.0} & 0.714\cellcolor{black!25.0} \\
\hline
InceptionResNet & 0.681\cellcolor{black!21.0} & 0.690\cellcolor{black!22.0} & 0.575\cellcolor{black!9.0} & 0.573\cellcolor{black!8.0} & 0.586\cellcolor{black!10.0} & 0.614\cellcolor{black!13.0} \\
\hline
MobileNet & 0.843\cellcolor{black!41.0} & 0.837\cellcolor{black!40.0} & 0.739\cellcolor{black!28.0} & 0.735\cellcolor{black!28.0} & 0.786\cellcolor{black!34.0} & 0.824\cellcolor{black!38.0} \\
\hline
MobileNet2 & 0.875\cellcolor{black!45.0} & 0.884\cellcolor{black!46.0} & 0.766\cellcolor{black!32.0} & 0.753\cellcolor{black!30.0} & 0.805\cellcolor{black!36.0} & 0.843\cellcolor{black!41.0} \\
\hline
DenseNet121 & 0.876\cellcolor{black!45.0} & 0.872\cellcolor{black!44.0} & 0.745\cellcolor{black!29.0} & 0.73\cellcolor{black!27.0} & 0.785\cellcolor{black!34.0} & 0.819\cellcolor{black!38.0} \\
\hline
DenseNet169 & 0.877\cellcolor{black!45.0} & 0.864\cellcolor{black!43.0} & 0.746\cellcolor{black!29.0} & 0.745\cellcolor{black!29.0} & 0.787\cellcolor{black!34.0} & 0.810\cellcolor{black!37.0} \\
\hline
DenseNet201 & 0.725\cellcolor{black!27.0} & 0.719\cellcolor{black!26.0} & 0.636\cellcolor{black!16.0} & 0.636\cellcolor{black!16.0} & 0.655\cellcolor{black!18.0} & 0.690\cellcolor{black!22.0} \\
\hline
NASNetMobile & 0.770\cellcolor{black!32.0} & 0.770\cellcolor{black!32.0} & 0.648\cellcolor{black!17.0} & 0.646\cellcolor{black!17.0} & 0.680\cellcolor{black!21.0} & 0.701\cellcolor{black!24.0} \\
\hline
Tuned ResNet & 0.912\cellcolor{black!49.0} & 0.906\cellcolor{black!48.0} & 0.845\cellcolor{black!41.0} & 0.835\cellcolor{black!40.0} & 0.865\cellcolor{black!43.0} & 0.878\cellcolor{black!45.0} \\
\hline
\end{tabular}
\end{minipage}
\end{scriptsize}
\end{table*}

\subsection{Detailed experiments with Pre-trained models} 
Given the success of the global method over the local method, we performed a more exhaustive set of tests on the global method, summarized for Ver$_1$ subset in Table ~\ref{tab:global_results_array}. 
Twelve different ImageNet classifiers were tested along with the fine-tuned ResNet50 network. 
For all networks, the output arrays from the last convolutional layer and the class output layer were saved. 
The classifiers given only the output layers have been omitted as they did not perform well compared to those given strictly the intermediate layer, or the intermediate and output layers concatenated. 
Then each combination of the previous two conditions was given to a classifier available in scikit-learn. 
With this common interface, over a dozen different classifiers with varying parameters were tested. 
Only a few are shown in Table~\ref{tab:global_results_array}. 
Based on these results, ResNet seemed to perform the best, which was consistent with the results in~\cite{transfer}.
Based on these results, and those from Ver$_2$ subset (not shown due to lack of space), the fine-tuned ResNet50 network had the best performance overall.


\section{Discussion and Future Work}

There are many additional methods which can be used as an add-on to the end-to-end trainable network. 
For example, face matching was one method we tested with mixed results.
On Ver$_1$ subset it gave a 1\% boost in AUC, while it had little effect on Ver$_2$ or Ver$_eval$ subset. 
Given that Ver$_1$ subset contained marathons and protests, while Ver$_2$ subset contained storms and airshows, this result was generally expected. 
Text extraction may be another area of interest as a supplementary decision. In visually inspecting the data, many images contained text which was unique to the event. 
This would include street signs, advertisements, and in the case of marathon events, race bibs. 
In future work we will attempt to accommodate images in their native resolutions. 
The dataset tested here consisted of images from 0.077 MP to 30MP. 
Resizing all images into the same height and/or width eliminates much of the additional information that large images contain. 
A brute force solution may be to sweep the feature extractor across multiple scales of an image pyramid. 
This will then give a pyramid of features extracted at different scales, leaving the open question of how to combine these. 
The two approaches represented here are a subset of this, where only the largest scales are used, and all others effectively ignored.
Finally, a collection of a larger dataset will facilitate 
in understanding how our methods will generalize in a real world application that may consist of hundreds of events and several thousand images. 


\section{Conclusions}

This paper demonstrated the viability of several possible methods for applying pre-trained ImageNet classifiers to the problem of event verification.
In particular, we explored analyzing an image at the global level and the local level, and studied the impact of fine-tuning a model for a specific dataset.
While the global classification methods out-performed the local methods in our experiments, more data will be needed to confirm this hypothesis.
The images included in our tested dataset contained less than a dozen events, and a broader range of data would clarify if there are certain types of events where one method may be superior. 
More research still needs to be done in this area, as it remains a pressing concern without a clear implementation at full scale.

\section{Acknowledgments} 

This research was developed with funding from the Defense Advanced Research Projects Agency (DARPA).
The views, opinions and/or findings expressed are those of the author and should not be interpreted as representing the official views or policies of the Department of Defense or the U.S. Government. 
The paper is approved for public release, distribution unlimited.

{\small
	\bibliographystyle{spiejour}
	\bibliography{event-ver.bbl}
}


\begin{biography}

\textbf{Michael Goebel} received his B.S. and M.S. degrees in Electrical Engineering from Binghamton University in 2016 and 2017. He is currently a PhD student in Electrical Engineering at University of California Santa Barbara.

\textbf{Arjuna Flenner} received his Ph.D. in Physics at the University of Missouri-Columbia located in Columbia MO in the year 2004. His major emphasis was mathematical Physics. After obtaining his Ph.D., Arjuna Flenner obtained a job as a research physicist for NAVAIR at China Lake CA. He won the 2013 Dr. Delores M. Etter Navy Scientist and Engineer award for his work on Machine Learning. 

\textbf{Lakshmanan Nataraj} received his B.E degree in Electronics and Communications Engineering from Sri Venkateswara College of Engineering, Anna University in 2007, and the Ph.D. degree in the Electrical and Computer Engineering from the University of California, Santa Barbara in 2015. 
He is currently a Senior Research Staff Member at Mayachitra Inc., Santa Barbara, CA. 
His research interests include malware analysis and image forensics.

\textbf{B. S. Manjunath} (F’05) received the Ph.D. degree in Electrical Engineering from the University of Southern California in 1991. He is currently a Distinguished Professor at the ECE Department at the University of California at Santa Barbara. 
He has co-authored about 300 peer-reviewed articles.
His current research interests include image processing, computer vision and biomedical image analysis.

\end{biography}


\section{Appendix}

\noindent Here we provide more experiments on Ver$_1$ and Ver$_2$ subsets (summarized in Tables.~\ref{tab:v1-inter},~\ref{tab:v1-out},~\ref{tab:v1-both},~\ref{tab:v2-inter},~\ref{tab:v2-out},~\ref{tab:v2-both}).
We test with more classifiers including Extra Trees classifiers (ET1, ET2), Nearest Neighbor classifier with different distance measures (L1,L2,Chebyshev), SVM on full feature and varying number of principal components (32,64,128,256), and a single layer dense neural network classifier trained for different lengths of time (exponential with respect to parameter shown in column header).

\vspace{-15pt}




















\begin{tiny}
\noindent
\begin{table*}
\centering
\caption{Ver$_1$: Intermediate Features Only}
\resizebox{\textwidth}{!}{
\begin{tabular}{|c||c|c|c|c|c|c|c|c|c|c|c|c|c|c|c|c|c|c|}
\hline
 & ET1 & ET2 & RF & 1NN L1 & 1NN Cheb & 1NN L2 & 2NN & 4NN & SVM & 32P+SVM & 64P+SVM & 128P+SVM & 256P+SVM & DNN4 & DNN3 & DNN2 & DNN1 & 128P+DNN\\
\hline
\hline
Xception & 0.766\cellcolor{black!32.0} & 0.764\cellcolor{black!31.0} & 0.753\cellcolor{black!30.0} & 0.641\cellcolor{black!17.0} & 0.581\cellcolor{black!9.0} & 0.638\cellcolor{black!16.0} & 0.675\cellcolor{black!21.0} & 0.695\cellcolor{black!23.0} & 0.617\cellcolor{black!14.0} & 0.465\cellcolor{black!0} & 0.460\cellcolor{black!0} & 0.456\cellcolor{black!0} & 0.451\cellcolor{black!0} & 0.709\cellcolor{black!25.0} & 0.701\cellcolor{black!24.0} & 0.697\cellcolor{black!23.0} & 0.652\cellcolor{black!18.0} & 0.740\cellcolor{black!28.0} \\
\hline
VGG16 & 0.880\cellcolor{black!45.0} & 0.877\cellcolor{black!45.0} & 0.874\cellcolor{black!44.0} & 0.751\cellcolor{black!30.0} & 0.678\cellcolor{black!21.0} & 0.775\cellcolor{black!33.0} & 0.818\cellcolor{black!38.0} & 0.845\cellcolor{black!41.0} & 0.409\cellcolor{black!0} & 0.415\cellcolor{black!0} & 0.417\cellcolor{black!0} & 0.421\cellcolor{black!0} & 0.419\cellcolor{black!0} & 0.782\cellcolor{black!33.0} & 0.782\cellcolor{black!33.0} & 0.788\cellcolor{black!34.0} & 0.703\cellcolor{black!24.0} & 0.846\cellcolor{black!41.0} \\
\hline
VGG19 & 0.883\cellcolor{black!46.0} & 0.883\cellcolor{black!46.0} & 0.874\cellcolor{black!44.0} & 0.756\cellcolor{black!30.0} & 0.641\cellcolor{black!17.0} & 0.768\cellcolor{black!32.0} & 0.820\cellcolor{black!38.0} & 0.838\cellcolor{black!40.0} & 0.404\cellcolor{black!0} & 0.414\cellcolor{black!0} & 0.415\cellcolor{black!0} & 0.416\cellcolor{black!0} & 0.415\cellcolor{black!0} & 0.785\cellcolor{black!34.0} & 0.795\cellcolor{black!35.0} & 0.782\cellcolor{black!33.0} & 0.712\cellcolor{black!25.0} & 0.826\cellcolor{black!39.0} \\
\hline
ResNet50 & 0.889\cellcolor{black!46.0} & 0.887\cellcolor{black!46.0} & 0.886\cellcolor{black!46.0} & 0.79\cellcolor{black!34.0} & 0.693\cellcolor{black!23.0} & 0.781\cellcolor{black!33.0} & 0.839\cellcolor{black!40.0} & 0.877\cellcolor{black!45.0} & 0.873\cellcolor{black!44.0} & 0.611\cellcolor{black!13.0} & 0.570\cellcolor{black!8.0} & 0.613\cellcolor{black!13.0} & 0.673\cellcolor{black!20.0} & 0.847\cellcolor{black!41.0} & 0.851\cellcolor{black!42.0} & 0.842\cellcolor{black!41.0} & 0.827\cellcolor{black!39.0} & 0.877\cellcolor{black!45.0} \\
\hline
InceptionV3 & 0.765\cellcolor{black!31.0} & 0.766\cellcolor{black!32.0} & 0.772\cellcolor{black!32.0} & 0.648\cellcolor{black!17.0} & 0.578\cellcolor{black!9.0} & 0.655\cellcolor{black!18.0} & 0.702\cellcolor{black!24.0} & 0.714\cellcolor{black!25.0} & 0.505\cellcolor{black!0.0} & 0.476\cellcolor{black!0} & 0.476\cellcolor{black!0} & 0.463\cellcolor{black!0} & 0.454\cellcolor{black!0} & 0.676\cellcolor{black!21.0} & 0.661\cellcolor{black!19.0} & 0.657\cellcolor{black!18.0} & 0.567\cellcolor{black!8.0} & 0.739\cellcolor{black!28.0} \\
\hline
InceptionResNet & 0.682\cellcolor{black!21.0} & 0.677\cellcolor{black!21.0} & 0.695\cellcolor{black!23.0} & 0.575\cellcolor{black!9.0} & 0.556\cellcolor{black!6.0} & 0.573\cellcolor{black!8.0} & 0.586\cellcolor{black!10.0} & 0.614\cellcolor{black!13.0} & 0.600\cellcolor{black!12.0} & 0.478\cellcolor{black!0} & 0.465\cellcolor{black!0} & 0.455\cellcolor{black!0} & 0.432\cellcolor{black!0} & 0.570\cellcolor{black!8.0} & 0.567\cellcolor{black!8.0} & 0.552\cellcolor{black!6.0} & 0.547\cellcolor{black!5.0} & 0.623\cellcolor{black!14.0} \\
\hline
MobileNet & 0.842\cellcolor{black!41.0} & 0.846\cellcolor{black!41.0} & 0.847\cellcolor{black!41.0} & 0.739\cellcolor{black!28.0} & 0.673\cellcolor{black!20.0} & 0.735\cellcolor{black!28.0} & 0.786\cellcolor{black!34.0} & 0.825\cellcolor{black!39.0} & 0.870\cellcolor{black!44.0} & 0.842\cellcolor{black!41.0} & 0.855\cellcolor{black!42.0} & 0.859\cellcolor{black!43.0} & 0.860\cellcolor{black!43.0} & 0.798\cellcolor{black!35.0} & 0.801\cellcolor{black!36.0} & 0.809\cellcolor{black!37.0} & 0.756\cellcolor{black!30.0} & 0.818\cellcolor{black!38.0} \\
\hline
MobileNet2 & 0.887\cellcolor{black!46.0} & 0.879\cellcolor{black!45.0} & 0.876\cellcolor{black!45.0} & 0.766\cellcolor{black!32.0} & 0.674\cellcolor{black!20.0} & 0.753\cellcolor{black!30.0} & 0.805\cellcolor{black!36.0} & 0.843\cellcolor{black!41.0} & 0.861\cellcolor{black!43.0} & 0.409\cellcolor{black!0} & 0.381\cellcolor{black!0} & 0.402\cellcolor{black!0} & 0.568\cellcolor{black!8.0} & 0.803\cellcolor{black!36.0} & 0.795\cellcolor{black!35.0} & 0.801\cellcolor{black!36.0} & 0.703\cellcolor{black!24.0} & 0.776\cellcolor{black!33.0} \\
\hline
DenseNet121 & 0.877\cellcolor{black!45.0} & 0.874\cellcolor{black!44.0} & 0.871\cellcolor{black!44.0} & 0.745\cellcolor{black!29.0} & 0.665\cellcolor{black!19.0} & 0.73\cellcolor{black!27.0} & 0.785\cellcolor{black!34.0} & 0.819\cellcolor{black!38.0} & 0.846\cellcolor{black!41.0} & 0.480\cellcolor{black!0} & 0.476\cellcolor{black!0} & 0.503\cellcolor{black!0.0} & 0.722\cellcolor{black!26.0} & 0.780\cellcolor{black!33.0} & 0.803\cellcolor{black!36.0} & 0.785\cellcolor{black!34.0} & 0.711\cellcolor{black!25.0} & 0.808\cellcolor{black!36.0} \\
\hline
DenseNet169 & 0.869\cellcolor{black!44.0} & 0.870\cellcolor{black!44.0} & 0.870\cellcolor{black!44.0} & 0.748\cellcolor{black!29.0} & 0.674\cellcolor{black!20.0} & 0.745\cellcolor{black!29.0} & 0.787\cellcolor{black!34.0} & 0.810\cellcolor{black!37.0} & 0.854\cellcolor{black!42.0} & 0.480\cellcolor{black!0} & 0.479\cellcolor{black!0} & 0.533\cellcolor{black!4.0} & 0.678\cellcolor{black!21.0} & 0.815\cellcolor{black!37.0} & 0.824\cellcolor{black!38.0} & 0.824\cellcolor{black!38.0} & 0.691\cellcolor{black!23.0} & 0.818\cellcolor{black!38.0} \\
\hline
DenseNet201 & 0.735\cellcolor{black!28.0} & 0.735\cellcolor{black!28.0} & 0.728\cellcolor{black!27.0} & 0.636\cellcolor{black!16.0} & 0.601\cellcolor{black!12.0} & 0.636\cellcolor{black!16.0} & 0.655\cellcolor{black!18.0} & 0.690\cellcolor{black!22.0} & 0.454\cellcolor{black!0} & 0.481\cellcolor{black!0} & 0.475\cellcolor{black!0} & 0.468\cellcolor{black!0} & 0.460\cellcolor{black!0} & 0.668\cellcolor{black!20.0} & 0.664\cellcolor{black!19.0} & 0.639\cellcolor{black!16.0} & 0.589\cellcolor{black!10.0} & 0.721\cellcolor{black!26.0} \\
\hline
NASNetMobile & 0.764\cellcolor{black!31.0} & 0.761\cellcolor{black!31.0} & 0.763\cellcolor{black!31.0} & 0.648\cellcolor{black!17.0} & 0.593\cellcolor{black!11.0} & 0.646\cellcolor{black!17.0} & 0.680\cellcolor{black!21.0} & 0.701\cellcolor{black!24.0} & 0.753\cellcolor{black!30.0} & 0.517\cellcolor{black!2.0} & 0.535\cellcolor{black!4.0} & 0.608\cellcolor{black!13.0} & 0.684\cellcolor{black!22.0} & 0.727\cellcolor{black!27.0} & 0.726\cellcolor{black!27.0} & 0.698\cellcolor{black!23.0} & 0.656\cellcolor{black!18.0} & 0.738\cellcolor{black!28.0} \\
\hline
Tuned ResNet & 0.909\cellcolor{black!49.0} & 0.907\cellcolor{black!48.0} & 0.905\cellcolor{black!48.0} & 0.845\cellcolor{black!41.0} & 0.775\cellcolor{black!33.0} & 0.831\cellcolor{black!39.0} & 0.867\cellcolor{black!44.0} & 0.876\cellcolor{black!45.0} & 0.902\cellcolor{black!48.0} & 0.704\cellcolor{black!24.0} & 0.555\cellcolor{black!6.0} & 0.519\cellcolor{black!2.0} & 0.547\cellcolor{black!5.0} & 0.897\cellcolor{black!47.0} & 0.896\cellcolor{black!47.0} & 0.896\cellcolor{black!47.0} & 0.896\cellcolor{black!47.0} & 0.836\cellcolor{black!40.0} \\
\hline
\end{tabular}
}
\label{tab:v1-inter}
\end{table*}

\noindent
\begin{table*}
\centering
\caption{Ver$_1$: Output Features Only}
\resizebox{\textwidth}{!}{
\begin{tabular}{|c||c|c|c|c|c|c|c|c|c|c|c|c|c|c|c|c|c|c|}
\hline
 & ET1 & ET2 & RF & 1NN L1 & 1NN Cheb & 1NN L2 & 2NN & 4NN & SVM & 32P+SVM & 64P+SVM & 128P+SVM & 256P+SVM & DNN4 & DNN3 & DNN2 & DNN1 & 128P+DNN\\
\hline
\hline
Xception & 0.567\cellcolor{black!8.0} & 0.572\cellcolor{black!8.0} & 0.607\cellcolor{black!12.0} & 0.54\cellcolor{black!4.0} & 0.55\cellcolor{black!6.0} & 0.56\cellcolor{black!7.0} & 0.567\cellcolor{black!8.0} & 0.579\cellcolor{black!9.0} & 0.531\cellcolor{black!3.0} & 0.572\cellcolor{black!8.0} & 0.564\cellcolor{black!7.0} & 0.560\cellcolor{black!7.0} & 0.555\cellcolor{black!6.0} & 0.586\cellcolor{black!10.0} & 0.587\cellcolor{black!10.0} & 0.585\cellcolor{black!10.0} & 0.525\cellcolor{black!3.0} & 0.586\cellcolor{black!10.0} \\
\hline
VGG16 & 0.869\cellcolor{black!44.0} & 0.866\cellcolor{black!43.0} & 0.868\cellcolor{black!44.0} & 0.693\cellcolor{black!23.0} & 0.625\cellcolor{black!15.0} & 0.635\cellcolor{black!16.0} & 0.689\cellcolor{black!22.0} & 0.727\cellcolor{black!27.0} & 0.532\cellcolor{black!3.0} & 0.762\cellcolor{black!31.0} & 0.773\cellcolor{black!32.0} & 0.763\cellcolor{black!31.0} & 0.748\cellcolor{black!29.0} & 0.801\cellcolor{black!36.0} & 0.816\cellcolor{black!37.0} & 0.825\cellcolor{black!39.0} & 0.619\cellcolor{black!14.0} & 0.807\cellcolor{black!36.0} \\
\hline
VGG19 & 0.879\cellcolor{black!45.0} & 0.879\cellcolor{black!45.0} & 0.873\cellcolor{black!44.0} & 0.695\cellcolor{black!23.0} & 0.646\cellcolor{black!17.0} & 0.646\cellcolor{black!17.0} & 0.708\cellcolor{black!24.0} & 0.742\cellcolor{black!29.0} & 0.525\cellcolor{black!3.0} & 0.800\cellcolor{black!36.0} & 0.799\cellcolor{black!35.0} & 0.792\cellcolor{black!35.0} & 0.757\cellcolor{black!30.0} & 0.822\cellcolor{black!38.0} & 0.834\cellcolor{black!40.0} & 0.830\cellcolor{black!39.0} & 0.647\cellcolor{black!17.0} & 0.810\cellcolor{black!37.0} \\
\hline
ResNet50 & 0.891\cellcolor{black!46.0} & 0.886\cellcolor{black!46.0} & 0.890\cellcolor{black!46.0} & 0.741\cellcolor{black!29.0} & 0.671\cellcolor{black!20.0} & 0.693\cellcolor{black!23.0} & 0.718\cellcolor{black!26.0} & 0.765\cellcolor{black!31.0} & 0.344\cellcolor{black!0} & 0.771\cellcolor{black!32.0} & 0.778\cellcolor{black!33.0} & 0.777\cellcolor{black!33.0} & 0.740\cellcolor{black!28.0} & 0.831\cellcolor{black!39.0} & 0.840\cellcolor{black!40.0} & 0.834\cellcolor{black!40.0} & 0.683\cellcolor{black!22.0} & 0.810\cellcolor{black!37.0} \\
\hline
InceptionV3 & 0.563\cellcolor{black!7.0} & 0.563\cellcolor{black!7.0} & 0.548\cellcolor{black!5.0} & 0.52\cellcolor{black!2.0} & 0.536\cellcolor{black!4.0} & 0.526\cellcolor{black!3.0} & 0.537\cellcolor{black!4.0} & 0.540\cellcolor{black!4.0} & 0.540\cellcolor{black!4.0} & 0.554\cellcolor{black!6.0} & 0.551\cellcolor{black!6.0} & 0.549\cellcolor{black!5.0} & 0.551\cellcolor{black!6.0} & 0.536\cellcolor{black!4.0} & 0.535\cellcolor{black!4.0} & 0.537\cellcolor{black!4.0} & 0.544\cellcolor{black!5.0} & 0.531\cellcolor{black!3.0} \\
\hline
InceptionResNet & 0.511\cellcolor{black!1.0} & 0.512\cellcolor{black!1.0} & 0.510\cellcolor{black!1.0} & 0.488\cellcolor{black!0} & 0.493\cellcolor{black!0} & 0.488\cellcolor{black!0} & 0.497\cellcolor{black!0} & 0.499\cellcolor{black!0} & 0.498\cellcolor{black!0} & nan\cellcolor{black!0} & nan\cellcolor{black!0} & nan\cellcolor{black!0} & nan\cellcolor{black!0} & 0.502\cellcolor{black!0.0} & 0.499\cellcolor{black!0} & 0.497\cellcolor{black!0} & 0.499\cellcolor{black!0} & nan\cellcolor{black!0} \\
\hline
MobileNet & 0.818\cellcolor{black!38.0} & 0.813\cellcolor{black!37.0} & 0.807\cellcolor{black!36.0} & 0.591\cellcolor{black!11.0} & 0.561\cellcolor{black!7.0} & 0.548\cellcolor{black!5.0} & 0.588\cellcolor{black!10.0} & 0.625\cellcolor{black!15.0} & 0.622\cellcolor{black!14.0} & 0.634\cellcolor{black!16.0} & 0.626\cellcolor{black!15.0} & 0.623\cellcolor{black!14.0} & 0.622\cellcolor{black!14.0} & 0.680\cellcolor{black!21.0} & 0.680\cellcolor{black!21.0} & 0.657\cellcolor{black!18.0} & 0.544\cellcolor{black!5.0} & 0.676\cellcolor{black!21.0} \\
\hline
MobileNet2 & 0.747\cellcolor{black!29.0} & 0.757\cellcolor{black!30.0} & 0.809\cellcolor{black!37.0} & 0.566\cellcolor{black!8.0} & 0.575\cellcolor{black!9.0} & 0.576\cellcolor{black!9.0} & 0.619\cellcolor{black!14.0} & 0.619\cellcolor{black!14.0} & 0.675\cellcolor{black!21.0} & 0.683\cellcolor{black!22.0} & 0.683\cellcolor{black!22.0} & 0.682\cellcolor{black!21.0} & 0.681\cellcolor{black!21.0} & 0.690\cellcolor{black!22.0} & 0.691\cellcolor{black!22.0} & 0.684\cellcolor{black!22.0} & 0.579\cellcolor{black!9.0} & 0.693\cellcolor{black!23.0} \\
\hline
DenseNet121 & 0.702\cellcolor{black!24.0} & 0.707\cellcolor{black!24.0} & 0.722\cellcolor{black!26.0} & 0.6\cellcolor{black!12.0} & 0.596\cellcolor{black!11.0} & 0.598\cellcolor{black!11.0} & 0.605\cellcolor{black!12.0} & 0.641\cellcolor{black!16.0} & 0.602\cellcolor{black!12.0} & 0.636\cellcolor{black!16.0} & 0.636\cellcolor{black!16.0} & 0.633\cellcolor{black!16.0} & 0.629\cellcolor{black!15.0} & 0.659\cellcolor{black!19.0} & 0.660\cellcolor{black!19.0} & 0.651\cellcolor{black!18.0} & 0.553\cellcolor{black!6.0} & 0.657\cellcolor{black!18.0} \\
\hline
DenseNet169 & 0.739\cellcolor{black!28.0} & 0.747\cellcolor{black!29.0} & 0.788\cellcolor{black!34.0} & 0.595\cellcolor{black!11.0} & 0.59\cellcolor{black!10.0} & 0.59\cellcolor{black!10.0} & 0.601\cellcolor{black!12.0} & 0.629\cellcolor{black!15.0} & 0.650\cellcolor{black!18.0} & 0.660\cellcolor{black!19.0} & 0.662\cellcolor{black!19.0} & 0.659\cellcolor{black!19.0} & 0.659\cellcolor{black!19.0} & 0.682\cellcolor{black!21.0} & 0.682\cellcolor{black!21.0} & 0.674\cellcolor{black!20.0} & 0.582\cellcolor{black!9.0} & 0.683\cellcolor{black!21.0} \\
\hline
DenseNet201 & 0.664\cellcolor{black!19.0} & 0.663\cellcolor{black!19.0} & 0.657\cellcolor{black!18.0} & 0.566\cellcolor{black!8.0} & 0.544\cellcolor{black!5.0} & 0.57\cellcolor{black!8.0} & 0.575\cellcolor{black!9.0} & 0.589\cellcolor{black!10.0} & 0.564\cellcolor{black!7.0} & 0.570\cellcolor{black!8.0} & 0.566\cellcolor{black!8.0} & 0.563\cellcolor{black!7.0} & 0.562\cellcolor{black!7.0} & 0.620\cellcolor{black!14.0} & 0.619\cellcolor{black!14.0} & 0.612\cellcolor{black!13.0} & 0.573\cellcolor{black!8.0} & 0.618\cellcolor{black!14.0} \\
\hline
NASNetMobile & 0.766\cellcolor{black!31.0} & 0.756\cellcolor{black!30.0} & 0.765\cellcolor{black!31.0} & 0.609\cellcolor{black!13.0} & 0.561\cellcolor{black!7.0} & 0.57\cellcolor{black!8.0} & 0.610\cellcolor{black!13.0} & 0.641\cellcolor{black!16.0} & 0.604\cellcolor{black!12.0} & 0.630\cellcolor{black!15.0} & 0.625\cellcolor{black!15.0} & 0.620\cellcolor{black!14.0} & 0.613\cellcolor{black!13.0} & 0.682\cellcolor{black!21.0} & 0.683\cellcolor{black!22.0} & 0.638\cellcolor{black!16.0} & 0.544\cellcolor{black!5.0} & 0.689\cellcolor{black!22.0} \\
\hline
Tuned ResNet & 0.910\cellcolor{black!49.0} & 0.915\cellcolor{black!49.0} & 0.911\cellcolor{black!49.0} & 0.831\cellcolor{black!39.0} & 0.813\cellcolor{black!37.0} & 0.846\cellcolor{black!41.0} & 0.864\cellcolor{black!43.0} & 0.875\cellcolor{black!45.0} & 0.924\cellcolor{black!50.0} & 0.912\cellcolor{black!49.0} & 0.913\cellcolor{black!49.0} & 0.912\cellcolor{black!49.0} & 0.914\cellcolor{black!49.0} & 0.913\cellcolor{black!49.0} & 0.914\cellcolor{black!49.0} & 0.913\cellcolor{black!49.0} & 0.915\cellcolor{black!49.0} & 0.903\cellcolor{black!48.0} \\
\hline
\end{tabular}
}
\label{tab:v1-out}
\end{table*}

\vspace{-5pt}

\noindent
\begin{table*}
\centering
\caption{Ver$_1$: Intermediate and Output Features}
\resizebox{\textwidth}{!}{
\begin{tabular}{|c||c|c|c|c|c|c|c|c|c|c|c|c|c|c|c|c|c|c|}
\hline
 & ET1 & ET2 & RF & 1NN L1 & 1NN Cheb & 1NN L2 & 2NN & 4NN & SVM & 32P+SVM & 64P+SVM & 128P+SVM & 256P+SVM & DNN4 & DNN3 & DNN2 & DNN1 & 128P+DNN\\
\hline
\hline
Xception & 0.760\cellcolor{black!31.0} & 0.757\cellcolor{black!30.0} & 0.756\cellcolor{black!30.0} & 0.641\cellcolor{black!17.0} & 0.581\cellcolor{black!9.0} & 0.638\cellcolor{black!16.0} & 0.675\cellcolor{black!21.0} & 0.695\cellcolor{black!23.0} & 0.650\cellcolor{black!18.0} & 0.466\cellcolor{black!0} & 0.462\cellcolor{black!0} & 0.454\cellcolor{black!0} & 0.452\cellcolor{black!0} & 0.721\cellcolor{black!26.0} & 0.718\cellcolor{black!26.0} & 0.731\cellcolor{black!27.0} & 0.696\cellcolor{black!23.0} & 0.738\cellcolor{black!28.0} \\
\hline
VGG16 & 0.879\cellcolor{black!45.0} & 0.882\cellcolor{black!45.0} & 0.876\cellcolor{black!45.0} & 0.751\cellcolor{black!30.0} & 0.678\cellcolor{black!21.0} & 0.775\cellcolor{black!33.0} & 0.818\cellcolor{black!38.0} & 0.845\cellcolor{black!41.0} & 0.446\cellcolor{black!0} & 0.415\cellcolor{black!0} & 0.421\cellcolor{black!0} & 0.421\cellcolor{black!0} & 0.419\cellcolor{black!0} & 0.788\cellcolor{black!34.0} & 0.786\cellcolor{black!34.0} & 0.794\cellcolor{black!35.0} & 0.779\cellcolor{black!33.0} & 0.835\cellcolor{black!40.0} \\
\hline
VGG19 & 0.884\cellcolor{black!46.0} & 0.885\cellcolor{black!46.0} & 0.884\cellcolor{black!46.0} & 0.756\cellcolor{black!30.0} & 0.641\cellcolor{black!17.0} & 0.768\cellcolor{black!32.0} & 0.820\cellcolor{black!38.0} & 0.838\cellcolor{black!40.0} & 0.594\cellcolor{black!11.0} & 0.412\cellcolor{black!0} & 0.413\cellcolor{black!0} & 0.415\cellcolor{black!0} & 0.417\cellcolor{black!0} & 0.808\cellcolor{black!37.0} & 0.810\cellcolor{black!37.0} & 0.776\cellcolor{black!33.0} & 0.741\cellcolor{black!28.0} & 0.810\cellcolor{black!37.0} \\
\hline
ResNet50 & 0.894\cellcolor{black!47.0} & 0.893\cellcolor{black!47.0} & 0.895\cellcolor{black!47.0} & 0.79\cellcolor{black!34.0} & 0.693\cellcolor{black!23.0} & 0.781\cellcolor{black!33.0} & 0.839\cellcolor{black!40.0} & 0.877\cellcolor{black!45.0} & 0.894\cellcolor{black!47.0} & 0.613\cellcolor{black!13.0} & 0.569\cellcolor{black!8.0} & 0.611\cellcolor{black!13.0} & 0.674\cellcolor{black!20.0} & 0.847\cellcolor{black!41.0} & 0.871\cellcolor{black!44.0} & 0.865\cellcolor{black!43.0} & 0.823\cellcolor{black!38.0} & 0.881\cellcolor{black!45.0} \\
\hline
InceptionV3 & 0.762\cellcolor{black!31.0} & 0.763\cellcolor{black!31.0} & 0.769\cellcolor{black!32.0} & 0.648\cellcolor{black!17.0} & 0.578\cellcolor{black!9.0} & 0.655\cellcolor{black!18.0} & 0.702\cellcolor{black!24.0} & 0.714\cellcolor{black!25.0} & 0.547\cellcolor{black!5.0} & 0.474\cellcolor{black!0} & 0.475\cellcolor{black!0} & 0.464\cellcolor{black!0} & 0.456\cellcolor{black!0} & 0.665\cellcolor{black!19.0} & 0.638\cellcolor{black!16.0} & 0.650\cellcolor{black!18.0} & 0.589\cellcolor{black!10.0} & 0.744\cellcolor{black!29.0} \\
\hline
InceptionResNet & 0.681\cellcolor{black!21.0} & 0.691\cellcolor{black!22.0} & 0.690\cellcolor{black!22.0} & 0.575\cellcolor{black!9.0} & 0.556\cellcolor{black!6.0} & 0.573\cellcolor{black!8.0} & 0.586\cellcolor{black!10.0} & 0.614\cellcolor{black!13.0} & 0.617\cellcolor{black!14.0} & 0.477\cellcolor{black!0} & 0.465\cellcolor{black!0} & 0.454\cellcolor{black!0} & 0.432\cellcolor{black!0} & 0.570\cellcolor{black!8.0} & 0.553\cellcolor{black!6.0} & 0.582\cellcolor{black!9.0} & 0.546\cellcolor{black!5.0} & 0.627\cellcolor{black!15.0} \\
\hline
MobileNet & 0.843\cellcolor{black!41.0} & 0.838\cellcolor{black!40.0} & 0.837\cellcolor{black!40.0} & 0.739\cellcolor{black!28.0} & 0.673\cellcolor{black!20.0} & 0.735\cellcolor{black!28.0} & 0.786\cellcolor{black!34.0} & 0.824\cellcolor{black!38.0} & 0.869\cellcolor{black!44.0} & 0.841\cellcolor{black!40.0} & 0.854\cellcolor{black!42.0} & 0.858\cellcolor{black!43.0} & 0.860\cellcolor{black!43.0} & 0.802\cellcolor{black!36.0} & 0.807\cellcolor{black!36.0} & 0.796\cellcolor{black!35.0} & 0.738\cellcolor{black!28.0} & 0.814\cellcolor{black!37.0} \\
\hline
MobileNet2 & 0.875\cellcolor{black!45.0} & 0.878\cellcolor{black!45.0} & 0.884\cellcolor{black!46.0} & 0.766\cellcolor{black!32.0} & 0.674\cellcolor{black!20.0} & 0.753\cellcolor{black!30.0} & 0.805\cellcolor{black!36.0} & 0.843\cellcolor{black!41.0} & 0.874\cellcolor{black!44.0} & 0.408\cellcolor{black!0} & 0.378\cellcolor{black!0} & 0.402\cellcolor{black!0} & 0.568\cellcolor{black!8.0} & 0.799\cellcolor{black!35.0} & 0.803\cellcolor{black!36.0} & 0.783\cellcolor{black!34.0} & 0.757\cellcolor{black!30.0} & 0.785\cellcolor{black!34.0} \\
\hline
DenseNet121 & 0.876\cellcolor{black!45.0} & 0.880\cellcolor{black!45.0} & 0.872\cellcolor{black!44.0} & 0.745\cellcolor{black!29.0} & 0.665\cellcolor{black!19.0} & 0.73\cellcolor{black!27.0} & 0.785\cellcolor{black!34.0} & 0.819\cellcolor{black!38.0} & 0.858\cellcolor{black!43.0} & 0.480\cellcolor{black!0} & 0.475\cellcolor{black!0} & 0.504\cellcolor{black!0.0} & 0.720\cellcolor{black!26.0} & 0.810\cellcolor{black!37.0} & 0.808\cellcolor{black!36.0} & 0.786\cellcolor{black!34.0} & 0.766\cellcolor{black!32.0} & 0.808\cellcolor{black!36.0} \\
\hline
DenseNet169 & 0.877\cellcolor{black!45.0} & 0.866\cellcolor{black!43.0} & 0.864\cellcolor{black!43.0} & 0.746\cellcolor{black!29.0} & 0.674\cellcolor{black!20.0} & 0.745\cellcolor{black!29.0} & 0.787\cellcolor{black!34.0} & 0.810\cellcolor{black!37.0} & 0.862\cellcolor{black!43.0} & 0.480\cellcolor{black!0} & 0.475\cellcolor{black!0} & 0.534\cellcolor{black!4.0} & 0.678\cellcolor{black!21.0} & 0.822\cellcolor{black!38.0} & 0.833\cellcolor{black!39.0} & 0.815\cellcolor{black!37.0} & 0.784\cellcolor{black!34.0} & 0.815\cellcolor{black!37.0} \\
\hline
DenseNet201 & 0.725\cellcolor{black!27.0} & 0.727\cellcolor{black!27.0} & 0.719\cellcolor{black!26.0} & 0.636\cellcolor{black!16.0} & 0.601\cellcolor{black!12.0} & 0.636\cellcolor{black!16.0} & 0.655\cellcolor{black!18.0} & 0.690\cellcolor{black!22.0} & 0.614\cellcolor{black!13.0} & 0.480\cellcolor{black!0} & 0.475\cellcolor{black!0} & 0.468\cellcolor{black!0} & 0.459\cellcolor{black!0} & 0.647\cellcolor{black!17.0} & 0.637\cellcolor{black!16.0} & 0.628\cellcolor{black!15.0} & 0.583\cellcolor{black!10.0} & 0.720\cellcolor{black!26.0} \\
\hline
NASNetMobile & 0.770\cellcolor{black!32.0} & 0.767\cellcolor{black!32.0} & 0.770\cellcolor{black!32.0} & 0.648\cellcolor{black!17.0} & 0.593\cellcolor{black!11.0} & 0.646\cellcolor{black!17.0} & 0.680\cellcolor{black!21.0} & 0.701\cellcolor{black!24.0} & 0.753\cellcolor{black!30.0} & 0.517\cellcolor{black!2.0} & 0.535\cellcolor{black!4.0} & 0.608\cellcolor{black!13.0} & 0.684\cellcolor{black!22.0} & 0.723\cellcolor{black!26.0} & 0.722\cellcolor{black!26.0} & 0.717\cellcolor{black!26.0} & 0.683\cellcolor{black!22.0} & 0.740\cellcolor{black!28.0} \\
\hline
Tuned ResNet & 0.912\cellcolor{black!49.0} & 0.914\cellcolor{black!49.0} & 0.906\cellcolor{black!48.0} & 0.845\cellcolor{black!41.0} & 0.775\cellcolor{black!33.0} & 0.835\cellcolor{black!40.0} & 0.865\cellcolor{black!43.0} & 0.878\cellcolor{black!45.0} & 0.913\cellcolor{black!49.0} & 0.701\cellcolor{black!24.0} & 0.548\cellcolor{black!5.0} & 0.523\cellcolor{black!2.0} & 0.550\cellcolor{black!6.0} & 0.905\cellcolor{black!48.0} & 0.902\cellcolor{black!48.0} & 0.901\cellcolor{black!48.0} & 0.903\cellcolor{black!48.0} & 0.851\cellcolor{black!42.0} \\
\hline
\end{tabular}
}
\label{tab:v1-both}
\end{table*}

\vspace{-5pt}

\noindent
\begin{table*}
\centering
\caption{Ver$_2$: Intermediate Features Only}
\resizebox{\textwidth}{!}{
\begin{tabular}{|c||c|c|c|c|c|c|c|c|c|c|c|c|c|c|c|c|c|c|}
\hline
 & ET1 & ET2 & RF & 1NN L1 & 1NN Cheb & 1NN L2 & 2NN & 4NN & SVM & 32P+SVM & 64P+SVM & 128P+SVM & 256P+SVM & DNN4 & DNN3 & DNN2 & DNN1 & 128P+DNN\\
\hline
\hline
Xception & 0.754\cellcolor{black!30.0} & 0.751\cellcolor{black!30.0} & 0.751\cellcolor{black!30.0} & 0.613\cellcolor{black!13.0} & 0.576\cellcolor{black!9.0} & 0.589\cellcolor{black!10.0} & 0.635\cellcolor{black!16.0} & 0.649\cellcolor{black!17.0} & 0.658\cellcolor{black!19.0} & 0.455\cellcolor{black!0} & 0.463\cellcolor{black!0} & 0.460\cellcolor{black!0} & 0.443\cellcolor{black!0} & 0.682\cellcolor{black!21.0} & 0.656\cellcolor{black!18.0} & 0.635\cellcolor{black!16.0} & 0.637\cellcolor{black!16.0} & 0.686\cellcolor{black!22.0} \\
\hline
VGG16 & 0.838\cellcolor{black!40.0} & 0.832\cellcolor{black!39.0} & 0.834\cellcolor{black!40.0} & 0.686\cellcolor{black!22.0} & 0.596\cellcolor{black!11.0} & 0.686\cellcolor{black!22.0} & 0.732\cellcolor{black!27.0} & 0.759\cellcolor{black!31.0} & 0.444\cellcolor{black!0} & 0.467\cellcolor{black!0} & 0.465\cellcolor{black!0} & 0.460\cellcolor{black!0} & 0.458\cellcolor{black!0} & 0.750\cellcolor{black!30.0} & 0.773\cellcolor{black!32.0} & 0.738\cellcolor{black!28.0} & 0.695\cellcolor{black!23.0} & 0.808\cellcolor{black!37.0} \\
\hline
VGG19 & 0.836\cellcolor{black!40.0} & 0.835\cellcolor{black!40.0} & 0.830\cellcolor{black!39.0} & 0.681\cellcolor{black!21.0} & 0.597\cellcolor{black!11.0} & 0.683\cellcolor{black!22.0} & 0.716\cellcolor{black!26.0} & 0.762\cellcolor{black!31.0} & 0.446\cellcolor{black!0} & 0.471\cellcolor{black!0} & 0.465\cellcolor{black!0} & 0.457\cellcolor{black!0} & 0.451\cellcolor{black!0} & 0.766\cellcolor{black!31.0} & 0.764\cellcolor{black!31.0} & 0.745\cellcolor{black!29.0} & 0.696\cellcolor{black!23.0} & 0.796\cellcolor{black!35.0} \\
\hline
ResNet50 & 0.857\cellcolor{black!42.0} & 0.848\cellcolor{black!41.0} & 0.856\cellcolor{black!42.0} & 0.681\cellcolor{black!21.0} & 0.628\cellcolor{black!15.0} & 0.673\cellcolor{black!20.0} & 0.716\cellcolor{black!26.0} & 0.754\cellcolor{black!30.0} & 0.827\cellcolor{black!39.0} & 0.432\cellcolor{black!0} & 0.435\cellcolor{black!0} & 0.403\cellcolor{black!0} & 0.387\cellcolor{black!0} & 0.813\cellcolor{black!37.0} & 0.818\cellcolor{black!38.0} & 0.832\cellcolor{black!39.0} & 0.791\cellcolor{black!34.0} & 0.864\cellcolor{black!43.0} \\
\hline
InceptionV3 & 0.742\cellcolor{black!29.0} & 0.740\cellcolor{black!28.0} & 0.751\cellcolor{black!30.0} & 0.589\cellcolor{black!10.0} & 0.584\cellcolor{black!10.0} & 0.594\cellcolor{black!11.0} & 0.634\cellcolor{black!16.0} & 0.673\cellcolor{black!20.0} & 0.527\cellcolor{black!3.0} & 0.481\cellcolor{black!0} & 0.478\cellcolor{black!0} & 0.474\cellcolor{black!0} & 0.472\cellcolor{black!0} & 0.684\cellcolor{black!22.0} & 0.649\cellcolor{black!17.0} & 0.649\cellcolor{black!17.0} & 0.644\cellcolor{black!17.0} & 0.715\cellcolor{black!25.0} \\
\hline
InceptionResNet & 0.685\cellcolor{black!22.0} & 0.676\cellcolor{black!21.0} & 0.689\cellcolor{black!22.0} & 0.542\cellcolor{black!5.0} & 0.552\cellcolor{black!6.0} & 0.546\cellcolor{black!5.0} & 0.595\cellcolor{black!11.0} & 0.610\cellcolor{black!13.0} & 0.638\cellcolor{black!16.0} & 0.448\cellcolor{black!0} & 0.442\cellcolor{black!0} & 0.437\cellcolor{black!0} & 0.602\cellcolor{black!12.0} & 0.563\cellcolor{black!7.0} & 0.553\cellcolor{black!6.0} & 0.536\cellcolor{black!4.0} & 0.502\cellcolor{black!0.0} & 0.633\cellcolor{black!16.0} \\
\hline
MobileNet & 0.731\cellcolor{black!27.0} & 0.723\cellcolor{black!26.0} & 0.722\cellcolor{black!26.0} & 0.626\cellcolor{black!15.0} & 0.582\cellcolor{black!9.0} & 0.636\cellcolor{black!16.0} & 0.653\cellcolor{black!18.0} & 0.679\cellcolor{black!21.0} & 0.757\cellcolor{black!30.0} & 0.686\cellcolor{black!22.0} & 0.710\cellcolor{black!25.0} & 0.727\cellcolor{black!27.0} & 0.733\cellcolor{black!28.0} & 0.737\cellcolor{black!28.0} & 0.743\cellcolor{black!29.0} & 0.732\cellcolor{black!27.0} & 0.692\cellcolor{black!23.0} & 0.745\cellcolor{black!29.0} \\
\hline
MobileNet2 & 0.762\cellcolor{black!31.0} & 0.760\cellcolor{black!31.0} & 0.764\cellcolor{black!31.0} & 0.636\cellcolor{black!16.0} & 0.576\cellcolor{black!9.0} & 0.621\cellcolor{black!14.0} & 0.660\cellcolor{black!19.0} & 0.670\cellcolor{black!20.0} & 0.716\cellcolor{black!25.0} & 0.448\cellcolor{black!0} & 0.446\cellcolor{black!0} & 0.436\cellcolor{black!0} & 0.433\cellcolor{black!0} & 0.701\cellcolor{black!24.0} & 0.728\cellcolor{black!27.0} & 0.701\cellcolor{black!24.0} & 0.651\cellcolor{black!18.0} & 0.711\cellcolor{black!25.0} \\
\hline
DenseNet121 & 0.767\cellcolor{black!32.0} & 0.758\cellcolor{black!30.0} & 0.769\cellcolor{black!32.0} & 0.629\cellcolor{black!15.0} & 0.597\cellcolor{black!11.0} & 0.633\cellcolor{black!15.0} & 0.648\cellcolor{black!17.0} & 0.685\cellcolor{black!22.0} & 0.724\cellcolor{black!26.0} & 0.452\cellcolor{black!0} & 0.453\cellcolor{black!0} & 0.465\cellcolor{black!0} & 0.550\cellcolor{black!6.0} & 0.698\cellcolor{black!23.0} & 0.714\cellcolor{black!25.0} & 0.701\cellcolor{black!24.0} & 0.696\cellcolor{black!23.0} & 0.703\cellcolor{black!24.0} \\
\hline
DenseNet169 & 0.766\cellcolor{black!32.0} & 0.765\cellcolor{black!31.0} & 0.778\cellcolor{black!33.0} & 0.634\cellcolor{black!16.0} & 0.574\cellcolor{black!8.0} & 0.628\cellcolor{black!15.0} & 0.649\cellcolor{black!17.0} & 0.674\cellcolor{black!20.0} & 0.735\cellcolor{black!28.0} & 0.443\cellcolor{black!0} & 0.452\cellcolor{black!0} & 0.495\cellcolor{black!0} & 0.550\cellcolor{black!6.0} & 0.708\cellcolor{black!25.0} & 0.707\cellcolor{black!24.0} & 0.723\cellcolor{black!26.0} & 0.681\cellcolor{black!21.0} & 0.729\cellcolor{black!27.0} \\
\hline
DenseNet201 & 0.755\cellcolor{black!30.0} & 0.756\cellcolor{black!30.0} & 0.754\cellcolor{black!30.0} & 0.601\cellcolor{black!12.0} & 0.579\cellcolor{black!9.0} & 0.599\cellcolor{black!11.0} & 0.646\cellcolor{black!17.0} & 0.688\cellcolor{black!22.0} & 0.437\cellcolor{black!0} & 0.487\cellcolor{black!0} & 0.485\cellcolor{black!0} & 0.478\cellcolor{black!0} & 0.468\cellcolor{black!0} & 0.664\cellcolor{black!19.0} & 0.654\cellcolor{black!18.0} & 0.667\cellcolor{black!20.0} & 0.657\cellcolor{black!18.0} & 0.704\cellcolor{black!24.0} \\
\hline
NASNetMobile & 0.726\cellcolor{black!27.0} & 0.725\cellcolor{black!27.0} & 0.728\cellcolor{black!27.0} & 0.613\cellcolor{black!13.0} & 0.586\cellcolor{black!10.0} & 0.589\cellcolor{black!10.0} & 0.632\cellcolor{black!15.0} & 0.658\cellcolor{black!18.0} & 0.718\cellcolor{black!26.0} & 0.620\cellcolor{black!14.0} & 0.638\cellcolor{black!16.0} & 0.665\cellcolor{black!19.0} & 0.692\cellcolor{black!23.0} & 0.669\cellcolor{black!20.0} & 0.673\cellcolor{black!20.0} & 0.674\cellcolor{black!20.0} & 0.615\cellcolor{black!13.0} & 0.684\cellcolor{black!22.0} \\
\hline
Tuned ResNet & 0.909\cellcolor{black!49.0} & 0.907\cellcolor{black!48.0} & 0.905\cellcolor{black!48.0} & 0.845\cellcolor{black!41.0} & 0.775\cellcolor{black!33.0} & 0.831\cellcolor{black!39.0} & 0.867\cellcolor{black!44.0} & 0.876\cellcolor{black!45.0} & 0.902\cellcolor{black!48.0} & 0.704\cellcolor{black!24.0} & 0.557\cellcolor{black!6.0} & 0.511\cellcolor{black!1.0} & 0.549\cellcolor{black!5.0} & 0.893\cellcolor{black!47.0} & 0.899\cellcolor{black!47.0} & 0.903\cellcolor{black!48.0} & 0.903\cellcolor{black!48.0} & 0.809\cellcolor{black!37.0} \\
\hline
\end{tabular}
}
\label{tab:v2-inter}
\end{table*}

\vspace{-5pt}

\noindent
\begin{table*}
\centering
\caption{Ver$_2$: Output Features Only}
\resizebox{\textwidth}{!}{
\begin{tabular}{|c||c|c|c|c|c|c|c|c|c|c|c|c|c|c|c|c|c|c|}
\hline
 & ET1 & ET2 & RF & 1NN L1 & 1NN Cheb & 1NN L2 & 2NN & 4NN & SVM & 32P+SVM & 64P+SVM & 128P+SVM & 256P+SVM & DNN4 & DNN3 & DNN2 & DNN1 & 128P+DNN\\
\hline
\hline
Xception & 0.616\cellcolor{black!13.0} & 0.619\cellcolor{black!14.0} & 0.618\cellcolor{black!14.0} & 0.530\cellcolor{black!3.0} & 0.537\cellcolor{black!4.0} & 0.530\cellcolor{black!3.0} & 0.567\cellcolor{black!8.0} & 0.608\cellcolor{black!13.0} & 0.644\cellcolor{black!17.0} & 0.641\cellcolor{black!17.0} & 0.640\cellcolor{black!16.0} & 0.646\cellcolor{black!17.0} & 0.649\cellcolor{black!17.0} & 0.636\cellcolor{black!16.0} & 0.637\cellcolor{black!16.0} & 0.642\cellcolor{black!17.0} & 0.535\cellcolor{black!4.0} & 0.641\cellcolor{black!16.0} \\
\hline
VGG16 & 0.833\cellcolor{black!40.0} & 0.838\cellcolor{black!40.0} & 0.840\cellcolor{black!40.0} & 0.643\cellcolor{black!17.0} & 0.629\cellcolor{black!15.0} & 0.616\cellcolor{black!13.0} & 0.672\cellcolor{black!20.0} & 0.710\cellcolor{black!25.0} & 0.788\cellcolor{black!34.0} & 0.793\cellcolor{black!35.0} & 0.793\cellcolor{black!35.0} & 0.797\cellcolor{black!35.0} & 0.799\cellcolor{black!35.0} & 0.772\cellcolor{black!32.0} & 0.782\cellcolor{black!33.0} & 0.794\cellcolor{black!35.0} & 0.630\cellcolor{black!15.0} & 0.771\cellcolor{black!32.0} \\
\hline
VGG19 & 0.831\cellcolor{black!39.0} & 0.833\cellcolor{black!39.0} & 0.823\cellcolor{black!38.0} & 0.638\cellcolor{black!16.0} & 0.603\cellcolor{black!12.0} & 0.619\cellcolor{black!14.0} & 0.671\cellcolor{black!20.0} & 0.733\cellcolor{black!28.0} & 0.790\cellcolor{black!34.0} & 0.808\cellcolor{black!37.0} & 0.814\cellcolor{black!37.0} & 0.810\cellcolor{black!37.0} & 0.802\cellcolor{black!36.0} & 0.794\cellcolor{black!35.0} & 0.809\cellcolor{black!37.0} & 0.818\cellcolor{black!38.0} & 0.609\cellcolor{black!13.0} & 0.787\cellcolor{black!34.0} \\
\hline
ResNet50 & 0.827\cellcolor{black!39.0} & 0.830\cellcolor{black!39.0} & 0.831\cellcolor{black!39.0} & 0.604\cellcolor{black!12.0} & 0.589\cellcolor{black!10.0} & 0.592\cellcolor{black!11.0} & 0.619\cellcolor{black!14.0} & 0.653\cellcolor{black!18.0} & 0.738\cellcolor{black!28.0} & 0.765\cellcolor{black!31.0} & 0.772\cellcolor{black!32.0} & 0.769\cellcolor{black!32.0} & 0.760\cellcolor{black!31.0} & 0.750\cellcolor{black!30.0} & 0.761\cellcolor{black!31.0} & 0.771\cellcolor{black!32.0} & 0.647\cellcolor{black!17.0} & 0.750\cellcolor{black!30.0} \\
\hline
InceptionV3 & 0.576\cellcolor{black!9.0} & 0.574\cellcolor{black!8.0} & 0.594\cellcolor{black!11.0} & 0.525\cellcolor{black!3.0} & 0.519\cellcolor{black!2.0} & 0.522\cellcolor{black!2.0} & 0.525\cellcolor{black!3.0} & 0.545\cellcolor{black!5.0} & 0.555\cellcolor{black!6.0} & 0.559\cellcolor{black!7.0} & 0.560\cellcolor{black!7.0} & 0.560\cellcolor{black!7.0} & 0.560\cellcolor{black!7.0} & 0.568\cellcolor{black!8.0} & 0.568\cellcolor{black!8.0} & 0.560\cellcolor{black!7.0} & 0.516\cellcolor{black!2.0} & 0.569\cellcolor{black!8.0} \\
\hline
InceptionResNet & 0.500\cellcolor{black!0.0} & 0.499\cellcolor{black!0} & 0.503\cellcolor{black!0.0} & 0.500\cellcolor{black!0.0} & 0.500\cellcolor{black!0.0} & 0.500\cellcolor{black!0.0} & 0.543\cellcolor{black!5.0} & 0.528\cellcolor{black!3.0} & 0.500\cellcolor{black!0.0} & nan\cellcolor{black!0} & nan\cellcolor{black!0} & nan\cellcolor{black!0} & nan\cellcolor{black!0} & 0.505\cellcolor{black!0.0} & 0.506\cellcolor{black!0.0} & 0.494\cellcolor{black!0} & 0.504\cellcolor{black!0.0} & nan\cellcolor{black!0} \\
\hline
MobileNet & 0.697\cellcolor{black!23.0} & 0.696\cellcolor{black!23.0} & 0.700\cellcolor{black!24.0} & 0.554\cellcolor{black!6.0} & 0.495\cellcolor{black!0} & 0.509\cellcolor{black!1.0} & 0.529\cellcolor{black!3.0} & 0.553\cellcolor{black!6.0} & 0.558\cellcolor{black!7.0} & 0.567\cellcolor{black!8.0} & 0.567\cellcolor{black!8.0} & 0.565\cellcolor{black!7.0} & 0.562\cellcolor{black!7.0} & 0.590\cellcolor{black!10.0} & 0.588\cellcolor{black!10.0} & 0.579\cellcolor{black!9.0} & 0.559\cellcolor{black!7.0} & 0.577\cellcolor{black!9.0} \\
\hline
MobileNet2 & 0.672\cellcolor{black!20.0} & 0.666\cellcolor{black!19.0} & 0.695\cellcolor{black!23.0} & 0.564\cellcolor{black!7.0} & 0.564\cellcolor{black!7.0} & 0.564\cellcolor{black!7.0} & 0.602\cellcolor{black!12.0} & 0.596\cellcolor{black!11.0} & 0.618\cellcolor{black!14.0} & 0.642\cellcolor{black!17.0} & 0.642\cellcolor{black!17.0} & 0.645\cellcolor{black!17.0} & 0.642\cellcolor{black!17.0} & 0.646\cellcolor{black!17.0} & 0.646\cellcolor{black!17.0} & 0.641\cellcolor{black!17.0} & 0.600\cellcolor{black!12.0} & 0.646\cellcolor{black!17.0} \\
\hline
DenseNet121 & 0.665\cellcolor{black!19.0} & 0.667\cellcolor{black!20.0} & 0.699\cellcolor{black!23.0} & 0.569\cellcolor{black!8.0} & 0.557\cellcolor{black!6.0} & 0.557\cellcolor{black!6.0} & 0.599\cellcolor{black!11.0} & 0.628\cellcolor{black!15.0} & 0.649\cellcolor{black!17.0} & 0.671\cellcolor{black!20.0} & 0.671\cellcolor{black!20.0} & 0.667\cellcolor{black!20.0} & 0.665\cellcolor{black!19.0} & 0.668\cellcolor{black!20.0} & 0.670\cellcolor{black!20.0} & 0.663\cellcolor{black!19.0} & 0.597\cellcolor{black!11.0} & 0.670\cellcolor{black!20.0} \\
\hline
DenseNet169 & 0.657\cellcolor{black!18.0} & 0.655\cellcolor{black!18.0} & 0.674\cellcolor{black!20.0} & 0.541\cellcolor{black!4.0} & 0.537\cellcolor{black!4.0} & 0.542\cellcolor{black!5.0} & 0.561\cellcolor{black!7.0} & 0.579\cellcolor{black!9.0} & 0.553\cellcolor{black!6.0} & 0.573\cellcolor{black!8.0} & 0.567\cellcolor{black!8.0} & 0.567\cellcolor{black!8.0} & 0.562\cellcolor{black!7.0} & 0.611\cellcolor{black!13.0} & 0.611\cellcolor{black!13.0} & 0.599\cellcolor{black!11.0} & 0.575\cellcolor{black!9.0} & 0.607\cellcolor{black!12.0} \\
\hline
DenseNet201 & 0.707\cellcolor{black!24.0} & 0.708\cellcolor{black!25.0} & 0.713\cellcolor{black!25.0} & 0.552\cellcolor{black!6.0} & 0.572\cellcolor{black!8.0} & 0.547\cellcolor{black!5.0} & 0.583\cellcolor{black!9.0} & 0.621\cellcolor{black!14.0} & 0.640\cellcolor{black!16.0} & 0.691\cellcolor{black!23.0} & 0.687\cellcolor{black!22.0} & 0.683\cellcolor{black!22.0} & 0.676\cellcolor{black!21.0} & 0.702\cellcolor{black!24.0} & 0.702\cellcolor{black!24.0} & 0.684\cellcolor{black!22.0} & 0.641\cellcolor{black!16.0} & 0.690\cellcolor{black!22.0} \\
\hline
NASNetMobile & 0.697\cellcolor{black!23.0} & 0.699\cellcolor{black!23.0} & 0.708\cellcolor{black!24.0} & 0.571\cellcolor{black!8.0} & 0.536\cellcolor{black!4.0} & 0.556\cellcolor{black!6.0} & 0.577\cellcolor{black!9.0} & 0.602\cellcolor{black!12.0} & 0.533\cellcolor{black!3.0} & 0.598\cellcolor{black!11.0} & 0.582\cellcolor{black!9.0} & 0.562\cellcolor{black!7.0} & 0.540\cellcolor{black!4.0} & 0.656\cellcolor{black!18.0} & 0.656\cellcolor{black!18.0} & 0.626\cellcolor{black!15.0} & 0.565\cellcolor{black!7.0} & 0.629\cellcolor{black!15.0} \\
\hline
Tuned ResNet & 0.910\cellcolor{black!49.0} & 0.915\cellcolor{black!49.0} & 0.911\cellcolor{black!49.0} & 0.831\cellcolor{black!39.0} & 0.813\cellcolor{black!37.0} & 0.846\cellcolor{black!41.0} & 0.864\cellcolor{black!43.0} & 0.875\cellcolor{black!45.0} & 0.924\cellcolor{black!50.0} & 0.912\cellcolor{black!49.0} & 0.913\cellcolor{black!49.0} & 0.913\cellcolor{black!49.0} & 0.914\cellcolor{black!49.0} & 0.916\cellcolor{black!50.0} & 0.914\cellcolor{black!49.0} & 0.913\cellcolor{black!49.0} & 0.910\cellcolor{black!49.0} & 0.910\cellcolor{black!49.0} \\
\hline
\end{tabular}
}
\label{tab:v2-out}
\end{table*}

\vspace{-5pt}

\noindent
\begin{table*}
\centering
\caption{Ver$_2$: Intermediate and Output Features}
\resizebox{\textwidth}{!}{
\begin{tabular}{|c||c|c|c|c|c|c|c|c|c|c|c|c|c|c|c|c|c|c|}
\hline
 & ET1 & ET2 & RF & 1NN L1 & 1NN Cheb & 1NN L2 & 2NN & 4NN & SVM & 32P+SVM & 64P+SVM & 128P+SVM & 256P+SVM & DNN4 & DNN3 & DNN2 & DNN1 & 128P+DNN\\
\hline
\hline
Xception & 0.743\cellcolor{black!29.0} & 0.742\cellcolor{black!29.0} & 0.746\cellcolor{black!29.0} & 0.613\cellcolor{black!13.0} & 0.576\cellcolor{black!9.0} & 0.589\cellcolor{black!10.0} & 0.635\cellcolor{black!16.0} & 0.649\cellcolor{black!17.0} & 0.671\cellcolor{black!20.0} & 0.457\cellcolor{black!0} & 0.465\cellcolor{black!0} & 0.459\cellcolor{black!0} & 0.443\cellcolor{black!0} & 0.681\cellcolor{black!21.0} & 0.710\cellcolor{black!25.0} & 0.656\cellcolor{black!18.0} & 0.642\cellcolor{black!17.0} & 0.686\cellcolor{black!22.0} \\
\hline
VGG16 & 0.838\cellcolor{black!40.0} & 0.847\cellcolor{black!41.0} & 0.841\cellcolor{black!41.0} & 0.686\cellcolor{black!22.0} & 0.596\cellcolor{black!11.0} & 0.686\cellcolor{black!22.0} & 0.732\cellcolor{black!27.0} & 0.759\cellcolor{black!31.0} & 0.479\cellcolor{black!0} & 0.468\cellcolor{black!0} & 0.464\cellcolor{black!0} & 0.460\cellcolor{black!0} & 0.457\cellcolor{black!0} & 0.763\cellcolor{black!31.0} & 0.774\cellcolor{black!32.0} & 0.749\cellcolor{black!29.0} & 0.757\cellcolor{black!30.0} & 0.829\cellcolor{black!39.0} \\
\hline
VGG19 & 0.834\cellcolor{black!40.0} & 0.837\cellcolor{black!40.0} & 0.840\cellcolor{black!40.0} & 0.680\cellcolor{black!21.0} & 0.597\cellcolor{black!11.0} & 0.683\cellcolor{black!22.0} & 0.716\cellcolor{black!26.0} & 0.762\cellcolor{black!31.0} & 0.431\cellcolor{black!0} & 0.470\cellcolor{black!0} & 0.466\cellcolor{black!0} & 0.457\cellcolor{black!0} & 0.450\cellcolor{black!0} & 0.775\cellcolor{black!33.0} & 0.761\cellcolor{black!31.0} & 0.752\cellcolor{black!30.0} & 0.699\cellcolor{black!23.0} & 0.791\cellcolor{black!35.0} \\
\hline
ResNet50 & 0.852\cellcolor{black!42.0} & 0.839\cellcolor{black!40.0} & 0.850\cellcolor{black!42.0} & 0.681\cellcolor{black!21.0} & 0.628\cellcolor{black!15.0} & 0.673\cellcolor{black!20.0} & 0.716\cellcolor{black!26.0} & 0.754\cellcolor{black!30.0} & 0.845\cellcolor{black!41.0} & 0.432\cellcolor{black!0} & 0.433\cellcolor{black!0} & 0.414\cellcolor{black!0} & 0.389\cellcolor{black!0} & 0.824\cellcolor{black!38.0} & 0.831\cellcolor{black!39.0} & 0.821\cellcolor{black!38.0} & 0.808\cellcolor{black!37.0} & 0.847\cellcolor{black!41.0} \\
\hline
InceptionV3 & 0.745\cellcolor{black!29.0} & 0.735\cellcolor{black!28.0} & 0.746\cellcolor{black!29.0} & 0.589\cellcolor{black!10.0} & 0.584\cellcolor{black!10.0} & 0.594\cellcolor{black!11.0} & 0.634\cellcolor{black!16.0} & 0.673\cellcolor{black!20.0} & 0.660\cellcolor{black!19.0} & 0.480\cellcolor{black!0} & 0.478\cellcolor{black!0} & 0.473\cellcolor{black!0} & 0.471\cellcolor{black!0} & 0.661\cellcolor{black!19.0} & 0.648\cellcolor{black!17.0} & 0.660\cellcolor{black!19.0} & 0.610\cellcolor{black!13.0} & 0.712\cellcolor{black!25.0} \\
\hline
InceptionResNet & 0.683\cellcolor{black!22.0} & 0.673\cellcolor{black!20.0} & 0.689\cellcolor{black!22.0} & 0.542\cellcolor{black!5.0} & 0.552\cellcolor{black!6.0} & 0.546\cellcolor{black!5.0} & 0.595\cellcolor{black!11.0} & 0.610\cellcolor{black!13.0} & 0.634\cellcolor{black!16.0} & 0.448\cellcolor{black!0} & 0.442\cellcolor{black!0} & 0.437\cellcolor{black!0} & 0.602\cellcolor{black!12.0} & 0.526\cellcolor{black!3.0} & 0.594\cellcolor{black!11.0} & 0.564\cellcolor{black!7.0} & 0.547\cellcolor{black!5.0} & 0.635\cellcolor{black!16.0} \\
\hline
MobileNet & 0.715\cellcolor{black!25.0} & 0.715\cellcolor{black!25.0} & 0.711\cellcolor{black!25.0} & 0.624\cellcolor{black!14.0} & 0.582\cellcolor{black!9.0} & 0.636\cellcolor{black!16.0} & 0.653\cellcolor{black!18.0} & 0.679\cellcolor{black!21.0} & 0.758\cellcolor{black!31.0} & 0.686\cellcolor{black!22.0} & 0.710\cellcolor{black!25.0} & 0.727\cellcolor{black!27.0} & 0.733\cellcolor{black!28.0} & 0.739\cellcolor{black!28.0} & 0.745\cellcolor{black!29.0} & 0.735\cellcolor{black!28.0} & 0.701\cellcolor{black!24.0} & 0.756\cellcolor{black!30.0} \\
\hline
MobileNet2 & 0.759\cellcolor{black!31.0} & 0.749\cellcolor{black!29.0} & 0.769\cellcolor{black!32.0} & 0.636\cellcolor{black!16.0} & 0.576\cellcolor{black!9.0} & 0.621\cellcolor{black!14.0} & 0.660\cellcolor{black!19.0} & 0.670\cellcolor{black!20.0} & 0.731\cellcolor{black!27.0} & 0.448\cellcolor{black!0} & 0.445\cellcolor{black!0} & 0.437\cellcolor{black!0} & 0.433\cellcolor{black!0} & 0.713\cellcolor{black!25.0} & 0.688\cellcolor{black!22.0} & 0.702\cellcolor{black!24.0} & 0.659\cellcolor{black!19.0} & 0.705\cellcolor{black!24.0} \\
\hline
DenseNet121 & 0.765\cellcolor{black!31.0} & 0.765\cellcolor{black!31.0} & 0.770\cellcolor{black!32.0} & 0.629\cellcolor{black!15.0} & 0.597\cellcolor{black!11.0} & 0.633\cellcolor{black!15.0} & 0.648\cellcolor{black!17.0} & 0.685\cellcolor{black!22.0} & 0.731\cellcolor{black!27.0} & 0.453\cellcolor{black!0} & 0.454\cellcolor{black!0} & 0.465\cellcolor{black!0} & 0.550\cellcolor{black!6.0} & 0.707\cellcolor{black!24.0} & 0.706\cellcolor{black!24.0} & 0.703\cellcolor{black!24.0} & 0.640\cellcolor{black!16.0} & 0.706\cellcolor{black!24.0} \\
\hline
DenseNet169 & 0.764\cellcolor{black!31.0} & 0.760\cellcolor{black!31.0} & 0.770\cellcolor{black!32.0} & 0.634\cellcolor{black!16.0} & 0.574\cellcolor{black!8.0} & 0.628\cellcolor{black!15.0} & 0.649\cellcolor{black!17.0} & 0.674\cellcolor{black!20.0} & 0.742\cellcolor{black!29.0} & 0.445\cellcolor{black!0} & 0.451\cellcolor{black!0} & 0.494\cellcolor{black!0} & 0.550\cellcolor{black!6.0} & 0.705\cellcolor{black!24.0} & 0.717\cellcolor{black!26.0} & 0.735\cellcolor{black!28.0} & 0.684\cellcolor{black!22.0} & 0.724\cellcolor{black!26.0} \\
\hline
DenseNet201 & 0.747\cellcolor{black!29.0} & 0.750\cellcolor{black!30.0} & 0.748\cellcolor{black!29.0} & 0.601\cellcolor{black!12.0} & 0.579\cellcolor{black!9.0} & 0.599\cellcolor{black!11.0} & 0.646\cellcolor{black!17.0} & 0.688\cellcolor{black!22.0} & 0.659\cellcolor{black!19.0} & 0.487\cellcolor{black!0} & 0.484\cellcolor{black!0} & 0.477\cellcolor{black!0} & 0.469\cellcolor{black!0} & 0.670\cellcolor{black!20.0} & 0.675\cellcolor{black!21.0} & 0.668\cellcolor{black!20.0} & 0.680\cellcolor{black!21.0} & 0.691\cellcolor{black!23.0} \\
\hline
NASNetMobile & 0.725\cellcolor{black!27.0} & 0.723\cellcolor{black!26.0} & 0.725\cellcolor{black!27.0} & 0.613\cellcolor{black!13.0} & 0.586\cellcolor{black!10.0} & 0.589\cellcolor{black!10.0} & 0.632\cellcolor{black!15.0} & 0.658\cellcolor{black!18.0} & 0.718\cellcolor{black!26.0} & 0.620\cellcolor{black!14.0} & 0.638\cellcolor{black!16.0} & 0.665\cellcolor{black!19.0} & 0.692\cellcolor{black!23.0} & 0.673\cellcolor{black!20.0} & 0.651\cellcolor{black!18.0} & 0.653\cellcolor{black!18.0} & 0.632\cellcolor{black!15.0} & 0.681\cellcolor{black!21.0} \\
\hline
Tuned ResNet & 0.912\cellcolor{black!49.0} & 0.914\cellcolor{black!49.0} & 0.906\cellcolor{black!48.0} & 0.845\cellcolor{black!41.0} & 0.775\cellcolor{black!33.0} & 0.835\cellcolor{black!40.0} & 0.865\cellcolor{black!43.0} & 0.878\cellcolor{black!45.0} & 0.913\cellcolor{black!49.0} & 0.702\cellcolor{black!24.0} & 0.555\cellcolor{black!6.0} & 0.524\cellcolor{black!2.0} & 0.545\cellcolor{black!5.0} & 0.901\cellcolor{black!48.0} & 0.899\cellcolor{black!47.0} & 0.904\cellcolor{black!48.0} & 0.892\cellcolor{black!47.0} & 0.836\cellcolor{black!40.0} \\
\hline
\end{tabular}
}
\label{tab:v2-both}
\end{table*}

\vspace{-5pt}



















\end{tiny}

\end{document}